\documentclass{article}
\usepackage{fullpage}
\usepackage{latexsym}
\usepackage[round]{natbib}
\usepackage{graphicx}
\usepackage[cmex10]{amsmath}
\usepackage{amssymb}
\usepackage{url}
\usepackage{booktabs}
\usepackage{color}
\usepackage{threeparttable}

\newcommand{\superscript}[1]{{$^{\textrm{#1}}$}}
\newcommand{\quotes}[1]{``#1''}
\definecolor{midblue}{rgb}{0.145,0.490,0.882}

\newcommand*{\eg}  {e.g.,\ }          
\newcommand*{\ie}  {i.e.,\ }          
\begin{document}
\title{Integer Programming Ensemble of Temporal Relations Classifiers}
\author{Catherine Kerr, Terri Hoare, Paula Carroll, Jakub Mare\v{c}ek}
\maketitle
\begin{abstract}
The extraction and understanding of temporal events and their relations are major challenges in natural language processing.
Processing text on a sentence-by-sentence or expression-by-expression basis often fails,
in part due to the challenge of capturing the global consistency of the text.
We present an ensemble method, which reconciles the outputs of multiple classifiers of temporal expressions
across the text using integer programming.
Computational experiments show that the ensemble improves upon the best individual results from two recent challenges,
SemEval-2013 TempEval-3 (Temporal Annotation) and SemEval-2016 Task 12 (Clinical TempEval).
\end{abstract}

\section{Introduction}\label{S.Intro}

The extraction of information from texts is of key importance in data mining. Applications range from web-scale tasks in search-engines \citep{32784,MovshovitzAttias2015}, machine reading in military intelligence \citep{bier2008}, to compiling situational awareness reports in disaster management, and distinguishing causes and effects in clinical applications \citep{Huang, styler2014temporal}.

Often, much is at stake: staff monitor an overload of text data sources such as print, digital, and broadcast media in order to compile situational awareness reports and coordinate responses to emergency situations. Much of this data is unstructured and monitoring activities are very costly, not only in terms of the direct costs, but also in terms of opportunity costs, e.g., putting more ``boots on the ground''. In clinical applications \citep{styler2014temporal}, understanding patients' medical history records may improve diagnostics and survival rates for individual patients, and may allow secondary research across all patient records, once the relative ordering of events can be recognised.


There are significant challenges to extracting useful information from texts. We often need to consider complete passages of text, because individual sentences may be ambiguous.
We also need to consider how to resolve conflicts, where coding schemes or identifiers in relational data may not be consistent \citep{Bhattacharya2007}.  \cite{Sauri2005} focus on the identification of temporal events in texts, while means of anaphora resolution across multiple sentences or multiple sources of data are addressed in \cite{Dong2005,li2011linking,dong2013data}. 
Based on multiple imperfect estimates of the meanings, possibly arrived at using multiple methods, we need to derive a single consistent estimate of the meaning of a passage of human natural language text.

\subsection{Temporal Reasoning Research Focus}
A series of natural language processing (NLP) competitions, known as the Temporal information extraction challenge (TempEval) \citep{Verhagen} and Semantic Evaluation (SemEval), have helped to focus NLP research on event and temporal relation processing. The competitions provide test text instances for computational experimentation, and specific NLP labelling tasks. The competitions use standardised data formats and performance measures so that competitors systems' performances on the tasks can be evaluated.

In this study, our particular focus is the temporal labelling of newsfeed documents of SemEval-2013 TempEval-3 Task C and clinical reports of TempEval 2016 Task 12. Given a set of test texts, TempEval-3 Task C challenges the competition participants to use their systems to link a set of events across newsfeed documents using a specified set of temporal relations. Clinical TempEval presents a set of clinical notes and pathology reports for cancer patients from the Mayo Clinic, and asks participants to extract temporal information and classify temporal relations. The test instances are annotated in a standard format, which is described in Section~\ref{S.NLP_problem}.

Our contribution is an integer-programming (IP) ensemble framework. Our framework synthesises concepts from a number of areas, which are introduced in Section~\ref{S.NLP_problem}. We describe our IP ensemble framework in Section \ref{S.IPensemble}. We generate multiple labellings of temporal events for the test instances by using several different classifiers provided by participants in the SemEval and TempEval challenges. The classifiers employ a variety of machine-learning techniques including support vector machines and methods inspired by maximum-entropy, which yield very different results. We use the IP ensemble model to reconcile the outputs of the classifiers and decide a final consistent labelling. The IP model maximises metrics associated with the estimates provided by the classifiers, such as the $F_1$ score, subject to a full set of consistency constraints. 

The ensemble allows the diversity of labellings to be exploited. It provides a single interpretation of the temporal relations that reconciles conflicts between the recommendations of the individual classifiers. The solution of the ensemble therefore provides a more consistent interpretation of the temporal information than the individual classifiers. Our results in Section~\ref{S.Results} show the IP ensemble yields an improvement over any of the individual classifiers who participated in either of the competitions.


\section{Temporal Reasoning in Natural Language Processing}\label{S.NLP_problem}
In order to automate temporal event analysis, we need to extract and analyse the temporal expressions and descriptions of events in textual reports and link the events appropriately. Consider the challenge to identify and order the temporal events in the following sample sentence from the TempEval-3 test data: 
\quotes{President Barack Obama arrived in refugee-flooded Jordan on Friday after scoring a diplomatic coup just before leaving Israel when Prime Minister Benjamin Netanyahu apologized to Turkey for a 2010 commando raid that killed nine activists on a Turkish vessel on a Gaza bound flotilla.}

We can identify several events:
Israel \emph{raided} a Turkish vessel,
\emph{killed} nine activists,
Benjamin Netanyahu \emph{apologized},
Barack Obama \emph{scored} a coup, \emph{left} Israel
for \emph{Jordan}.
Further, we can identify that the events happened in this order, possibly with leaving Israel happening at the
same time as entering Jordan, and also possibly with the apology and the scoring of the diplomatic coup coinciding.
We can also identify two temporal expressions (2010, Friday) and associate them with the respective events.
While this may be trivial for humans, it is not trivial for a computer.
Indeed, it may not even be clear how to formalise the problem.

Over the past half century, several frameworks have been designed to describe both qualitative and quantitative relations between time points and time intervals, as well as combinations of these \citep{Daykin}. A calculus, or relational algebra, for ordering the relationship of temporal events gives us a way to formalise a set of (temporal) rules. The application of the global rules of such a calculus
on the full set of temporal relations provides a necessary and sufficient condition for consistency across the relations in a news-feed document \citep{Ladkin}, and allows us to produce meaningful results for the temporal reasoning problem.
In the following, we describe one such algebra and show how it can be used in temporal reasoning
with the support of a temporal markup language and graphical models.

\subsection{Allen's Interval Algebra}\label {S.AllenAlg}
The interval algebra of \cite{Allen} is a popular formalism for reasoning about the relationships between time intervals. \cite{Allen} provides an algebra of binary relations on intervals where Time is represented by intervals and there are thirteen mutually exclusive dyadic relations between pairs of intervals. These thirteen relation types (\textit{reltype}s) are Precedes, Meets, Overlaps, Starts, Finishes, and During, their converses (Preceded by, Met by, Overlapped by, Started by, Finished by and Contains), and the equals relation. The thirteen basic reltypes precisely characterise the relative start and end points of two temporal intervals.

Allen also describes the concepts of temporal path consistency and transitive closure of the network of relations
between intervals, 
concepts which were extended by \cite{Ladkin}. Between any two temporal intervals exactly one of the relations holds. Transitivity properties are used to determine which relations may hold between pairs of intervals based on known relations between other intervals.
The problem of deciding which reltype holds can be represented as a directed graph of the interval algebra. Paths are consistent if there are no impossible relations among all triples of nodes in the graph.

\begin{figure}[http!]
\begin{center}
 \includegraphics[width=0.7\linewidth]{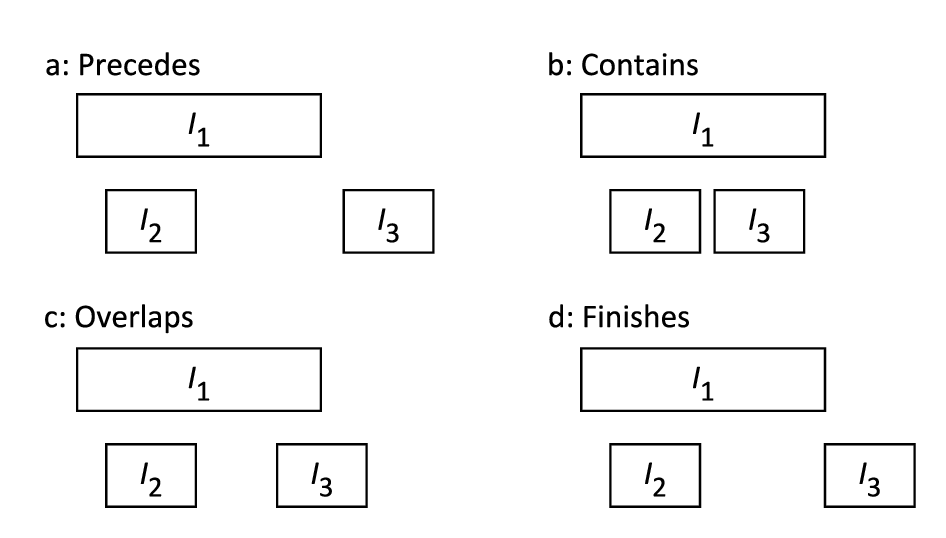}
\caption{An example of possible classifications of reltypes between $I_1$ and $I_3$, considering $I_1$ contains $I_2$, and $I_2$ precedes $I_3$.}
    \label{fig:AllenIA}
    \end{center}
\end{figure}

\begin{figure}[http!]
	\begin{center}
        \includegraphics[width=1.0\linewidth]{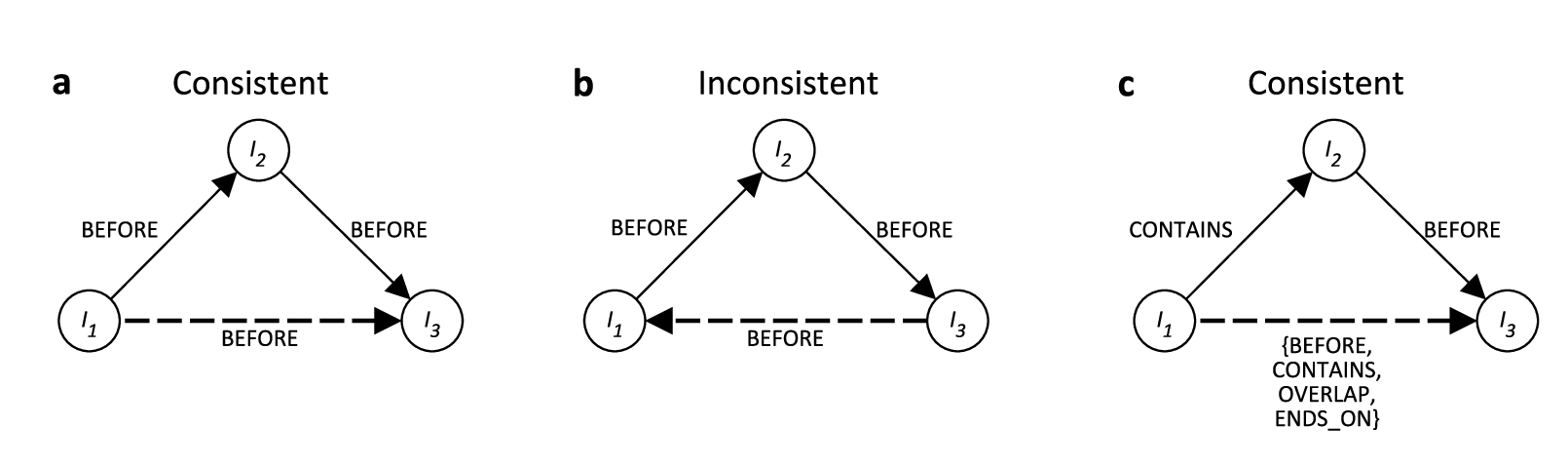}
	     	\caption{Examples of reasoning with Allen's Interval Algebra.}
		\label{fig:AllenEx}
	\end{center}
\end{figure}

Consider a simple example, where $I_1$ contains $I_2$, and $I_2$ precedes $I_3$. What can be said about the relationship type (reltype) between $I_1$ and $I_3$?
Figure \ref{fig:AllenIA} illustrates the four options
of completing the transitive closure of the interval network
within Allen's Interval Algebra, with time on the horizontal axis in each subfigure.
In Figure~\ref{fig:AllenEx}, we illustrate the inference.
Solid arcs show the known reltype, while dashed arcs represent the reltype we are attempting to infer.
In Subfigure~\ref{fig:AllenEx}.a, the reltype Precedes is consistent, while in Subfigure~\ref{fig:AllenEx}.b, it is not.
Subfigure~\ref{fig:AllenEx}.c suggests that we must decide, which of the reltypes is the most appropriate labelling of the dashed arc.

\cite{Allen} describes a polynomial time constraint propagation algorithm to compute an approximation of the strongest implied relation for each pair of intervals. He demonstrates that algorithm is correct and never infers an inconsistent path between a pair of intervals. However, Allen also shows that the algorithm is incomplete and does not make all the inferences, \ie complete the transitive closure. \cite{nebel1995reasoning} discuss the computational challenge and tractability of algorithms for the transitive closure of interval algebra networks. They note that the problem is NP-Complete in the full algebra, but propose polynomial time algorithms for subsets  of the algebra.

Much of the recent interest in Allen's temporal intervals is due to the TempEval competitions.
In particular, SemEval-2013 TempEval-3 Task C (Temporal Annotation) and SemEval-2016 Task 12 (Clinical TempEval) are recent competitions, which use two different subsets of Allen's interval algebra.
We describe
the competitions in further detail in Section~\ref{S.NLPCompetitions},
including the temporal markup language, TimeML, which the competitions use to encode the Allen intervals, and which we describe in Section~\ref{S.TimeML},
after we present the two subsets of Allen's interval algebra used in the two competitions.

Allen encoded transitivity information in a
composition table, so that reasoning in the interval calculus starts with a look-up in the table.
Tables~\ref{tab:allenalgebra} and \ref{tab:intalg} show the relations composition tables for the two competitions,
in the TimeML format of the reltype attributes. 
The left-most columns 
show the reltypes alongside a short code, \eg 'p' for BEFORE (which is the TimeML encoding of Precedes in Allen's Interval Algrbra). Rows represent $I_1 \circ I_2$ and columns represent $I_2 \circ I_3$, where $\circ$ represents any of the reltypes. The cell where the row and column intersect lists possible reltypes of $I_1 \circ I_3$.
For example, given that interval $I_1$ is before $I_2$ and that $I_2$ is before $I_3$, we can infer from Table~\ref{tab:allenalgebra} that $I_1$ is before $I_3$ as in Subfigure~\ref{fig:AllenEx}.a. This also tells us that $I_3$ before $I_1$ is inconsistent with the given information, as we saw in Subfigure~\ref{fig:AllenEx}.b.

Similarly, the reltypes used in the SemEval-2016 clinical text challenge are shown in Table~\ref{tab:intalg}. Taking as an example $I_1 \circ I_2$ is $c$ (Contains) and $I_2 \circ I_3$ is $p$ (Before), as we saw in Subfigure~\ref{fig:AllenEx}.c, then $I_1 \circ I_2 \circ I_3 = I_1 \circ I_3 \in \{$p, c, o, bi$\}$, \ie the composite relation is one of Before ($p$), Contains ($c$), Overlap ($o$), or Ends-on ($bi$, which is the TimeML attribute of Allen's Finishes).

\begin{table}[http!]
  \centering
  \caption{
  Transitivity relations in SemEval-2013 TempEval-3 Task C} 
  \begin{tabular}{p{1.6cm}p{0.4cm}p{0.4cm}p{0.4cm}p{0.4cm}p{0.4cm}p{0.4cm}p{0.4cm}p{0.4cm}p{0.4cm}p{0.4cm}p{0.4cm}p{0.4cm}}
    \toprule
    & p & pi & o & oi & I & Ii & m & mi & s & si & f & fi \\
    \midrule
  BEFORE (p) & p & . & p & p, d, o, m, s & p & p & p & p, d, o, m, s & p & p & p, d, o, m, s & p \\
  \hline
  AFTER (pi) & . & pi & pi, d, oi, mi, f & pi & pi & pi & pi, d, oi, mi, f & pi & pi, d, oi, mi, f & pi & pi & pi \\
  \hline
  INCLUDES (o) & p & pi, di, oi, mi, si & p, o, m & . & o & o & p & di, oi, si & o & di, o, fi & d, o, s & p, o, m \\
  \hline
  IS\_INCLUDED (oi) & p, di, o, m, fi & pi & . & pi, oi, mi & oi & oi & di, o, fi & pi & di, o, fi & pi, oi, mi & oi & di, oi, si \\
  \hline
  IDENTITY (I) & p & pi & o & oi & I & I & m & mi & s & si & f & fi \\
  \hline
  SIMULTAN-EOUS (Ii) & p & pi & o & oi & Ii & Ii & m & mi & s & si & f & fi \\
  \hline
  IBEFORE (m) & p & pi, di, oi, mi, si & p & d, o, s & m & m & p & I, Ii, f, fi & m & m & d, o, s & p \\
  \hline
  IAFTER (mi) & p, di, o, m, fi & pi & d, oi, f & pi & mi & mi & I, Ii, s, si & pi & d, oi, f & pi & mi & mi \\
  \hline
  BEGINS (s) & p & pi & p, o, m & d, oi, f & s & s & p & mi & s & I, Ii, s, si & . & p, o, m \\
  \hline
  BEGUN\_BY (si) & p, di, o, m, fi & pi & di, o, fi & oi & si & si & di, o, fi & mi & I, Ii, s, si & si & oi & di \\
  \hline
  ENDS (f) & p & pi & d, o, s & pi, oi, mi & d & f & m & pi & . & pi, oi, mi & d & I, Ii, f, fi \\
  \hline
  ENDED\_BY (fi) & p & pi, di, oi, mi, si & o & di, oi, si & fi & fi & m & di, oi, si & o & di & I, Ii, f, fi & fi \\
    \bottomrule
    \end{tabular}%
  \label{tab:allenalgebra}%
\end{table}%

\begin{table}[http!]
  \centering
  \caption{Transitivity relations in TempEval 2016 Task 12.}
    \begin{tabular}{p{2.8cm}p{0.7cm}p{0.7cm}p{0.7cm}p{0.7cm}p{0.7cm}p{0.7cm}p{0.7cm}}
    \toprule
    Relation & p      & pi    & c        & ci           & o       & b         & bi \\
    \midrule
	BEFORE (p) & p & . & p & p, ci, o, bi & p, ci, o, bi & p, ci, o, bi & p \\
	\hline
	AFTER (pi) & . & pi & pi & pi, ci, o, b & pi, ci, o, b & pi, ci, o, b & pi \\
	\hline
	CONTAINS (c) & p, c, o, bi & pi, c, o, b & c & c, ci, o & c, o & c, o & c, o \\
	\hline
	CONTAINS\_INV (ci) & p & pi & . & ci & p, pi, o, b, bi & pi & p \\
	\hline
	OVERLAP (o) & p, c, o, bi & pi, c, o, b & c, o & ci, o & ci, o, b, bi & pi, c, o & p, c, o \\
	\hline
	BEGINS\_ON (b) & p, c, o, bi & pi & pi & ci, o & pi, ci, o & pi & c, o \\
	\hline
	ENDS\_ON (bi) & p & pi, c, o, b & p & ci, o & p, ci, o & c, o & p \\
    \bottomrule
    \end{tabular}%
  \label{tab:intalg}%
\end{table}%

\subsection{TimeML Markup Language}\label{S.TimeML}
Temporal relations and events in text are tagged for the TempEval competitions using the TimeML markup language so that Allen's algebra can be used. TimeML is a temporal information standard markup language for labelling events, times, and their temporal relations in a text document \citep{Pustejovsky2005545}.
TimeML uses four primary tag types: \emph{EVENT} for temporal events, \emph{TIMEX3} for temporal expressions, \emph{SIGNAL} for temporal signals, and \emph{TLINK} to represent relationships between events and times or between events  \citep{Sauri}. 
TimeML encodes the Allen Intervals and includes a fourteenth relation to include the converse of equals. Each event has a unique identifier (EVENT-ID) and is associated (by a TLINK) with some time expression (TIMEX).

\cite{styler2014temporal} propose an extension to capture the requirements for the annotation of temporal information within clinical narratives.

In the example of the previous subsection, the event of \quotes{President Obama's arrival} is associated with the time expression of \quotes{Friday} by means of a TLINK. 
When we use $E^i$ ($T_i$) to denote the $i^{th}$ event (time expression, respectively),
the full TempEval-3 annotation of the sentence can be:

\noindent
President Barack Obama
\emph{arrived}\superscript{E1} in refugee-flooded Jordan on
\emph{Friday}\superscript{T1} after
\emph{scoring}\superscript{E2} a diplomatic
\emph{coup}\superscript{E3} just before
\emph{leaving}\superscript{E4} Israel when Prime Minister Benjamin Netanyahu
\emph{apologized}\superscript{E5} to Turkey for a
\emph{2010}\superscript{T2} commando
\emph{raid}\superscript{E6} that
\emph{killed}\superscript{E7} nine activists on a Turkish
 vessel on a Gaza bound flotilla.

Our interpretation of the temporal ordering of these events is:

\noindent
E2 (scoring),
E3 (coup) and
E5 (apologized) are
SIMULTANEOUS events. They occur
BEFORE
E4 (leaving),  which occurs
BEFORE
E1 (arrived).
E1 \ IS\_INCLUDED in
T1 (Friday).
E6 (raid) IS\_INCLUDED in
T2 (2010) and
E7 (killed) IS\_INCLUDED in
E6. Events
E1,
E2,
E3,
E4, 
E5 occurred
AFTER T2 (2010).

However, it is not clear if
E2,
E3 and
E5 occurred on or before
T1 (Friday).

\subsection{Event Graphs}\label {S.EGs}
\begin{figure}[http!]
	\begin{center}
        \includegraphics[width=1.0\linewidth,clip,trim=0 10cm 0 0]{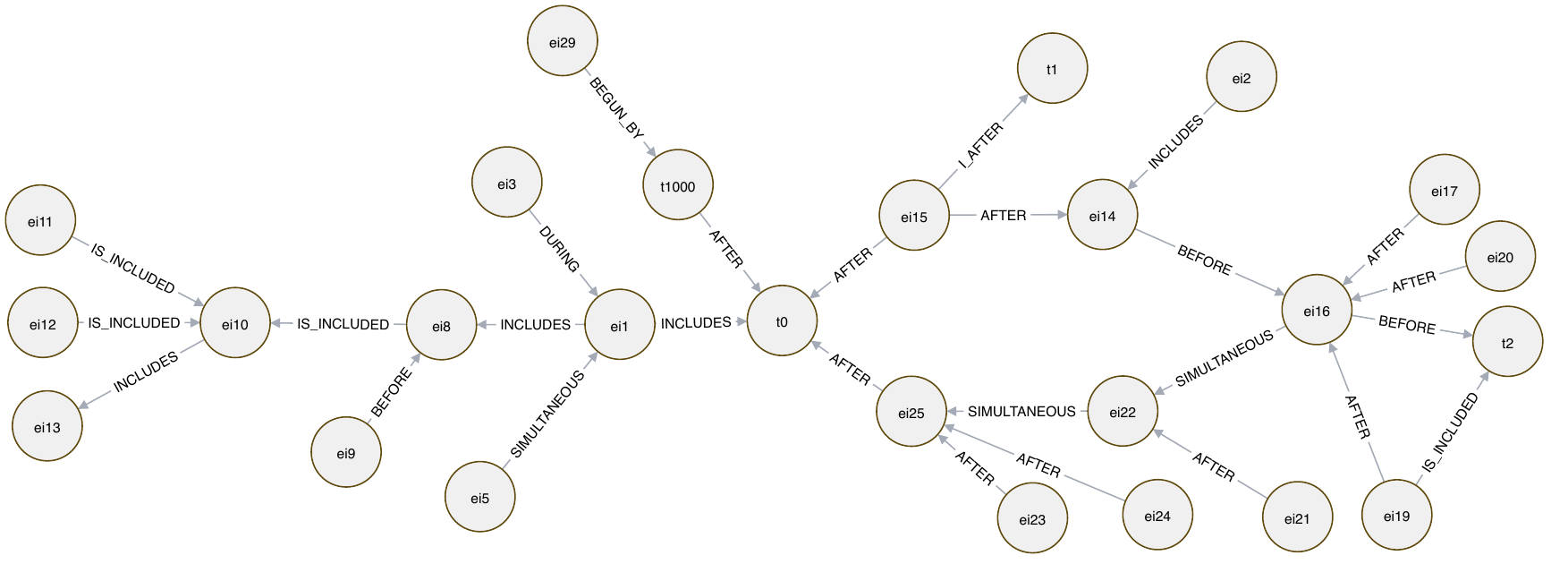}
	    	\caption{Event graph of hand annotated news feed (bbc\_20130322\_1600).}
		\label{fig:Platinum}
	\end{center}
\end{figure}

Event graphs (EG) are a useful way to represent events \citep{Glavas}. An event graph is a labeled graph in which nodes represent individual events, and edges represent semantic relations between events (\eg temporal relations). We can adapt EGs to model TimeML annotated text where nodes are the union of events and times and arcs are the TimeML TLINKs with arc weights representing the values of any of the fourteen types of relations or NONE (denoting no relation). 
The manual annotation of events, time expressions, and their temporal relations by experts provides a \quotes{Platinum} standard benchmark against which competitors' systems can be evaluated.
Figure~\ref{fig:Platinum} is 
an example of an EG of a human annotated BBC news feed, which was used in the TempEval 2013 challenge and starts with:


\noindent
Heavy \emph{snow}\superscript{E1}
is  \emph{causing}\superscript{E2}
 \emph{disruption}\superscript{E3}
to transport across the UK, with heavy  \emph{rainfall}\superscript{E1000}
 \emph{bringing}\superscript{E5}
 \emph{flooding}\superscript{E1001}
to the south-west of England. 
Rescuers  \emph{searching}\superscript{E6}
for a woman 
 \emph{trapped}\superscript{E7}
in a landslide at her home in Looe, Cornwall,  \emph{said}\superscript{E8}
they had  \emph{found}\superscript{E9}
a body. 

\begin{figure}[htbp]
	\begin{center}
		\includegraphics[width=0.75\linewidth]{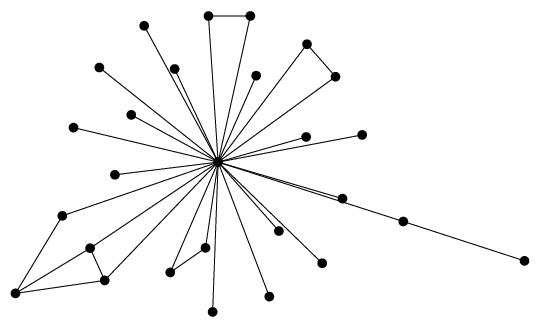} 
	    	\caption{EG of news feed (bbc\_20130322\_1600) annotated by Cleartk-2 \citep{Bethard2013}.}
		\label{fig:cleartk2}
	\end{center}
	\begin{center}
		\includegraphics[width=0.75\linewidth]{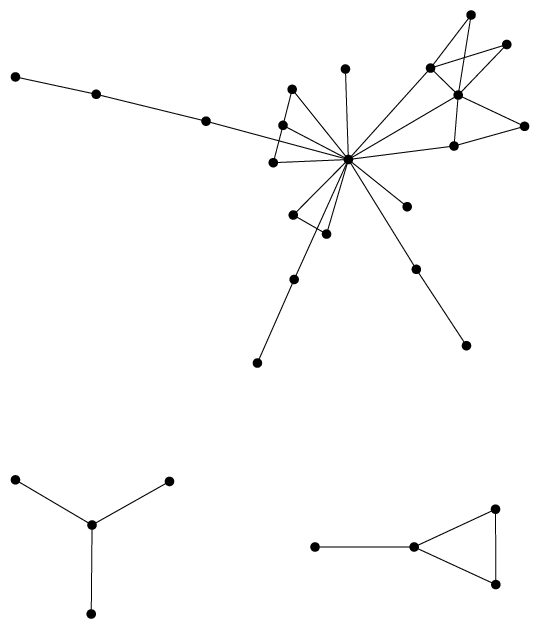} 
	    	\caption{EG of news feed (bbc\_20130322\_1600) annotated by
             Navytime-1 \citep{Chambers2013}.
            }
		\label{fig:navytime1}
	\end{center}
\end{figure}

For comparison, Figures~\ref{fig:cleartk2} and  \ref{fig:navytime1} show the EGs produced by two participants in SemEval-2013 TempEval-3 Task C: Cleartk-2 \citep{Bethard2013} and Navytime-1 \citep{Chambers2013} for the same BBC news feed in Figure~\ref{fig:Platinum}. Edge weights are omitted for clarity of the illustration.


The EGs demonstrate the diversity between the classifier outputs and the challenge in reconciling conflicting recommendations from the individual classifiers. 

\subsection{Evaluating the Performance of Classifiers}\label{S.NLPCompetitions}
The TempEval and SemEval challenges focus on the detection and classification of temporal relations between events. SemEval-2013 TempEval-3 participants trained their classifiers on a training data set and tested them on a platinum dataset of twenty newsfeeds \citep{UzZaman2012}. SemEval-2013 TempEval-3 Task C considered  all fourteen TimeML temporal relations \citep{UzZaman2012Th}. 

Clinical TempEval was introduced in SemEval-2016. The focus of TempEval 2016 Task 12 was clinical data (i.e., medical records) and six out of the fourteen relations of Allen's interval algebra. Participant systems were trained and evaluated on a corpus of clinical and pathology notes from the Mayo Clinic, annotated with an extension of TimeML for the clinical domain.
We are interested in Phase 2 of the competition, where participants were provided with manually annotated EVENTs and TIMEX3s, annotated by the THYME project (\texttt{thyme.healthnlp.org}).
The participating classifiers used the same baseline data to extract temporal relations between known events and time expressions. Phase 2 was further subdivided into two parts: First, to
 identify relations between events and the document creation time (DOCTIMEREL); second, to identify narrative container relations (CONTAINS).
A narrative container can be thought of as a temporal bucket into which an EVENT or series of EVENTs may fall \citep{pustejovsky2011increasing}. These narrative containers are often represented (or ``anchored'') by dates or other temporal expressions (within which a variety of different EVENTs occur).
The objective is to link EVENTs to narrative containers and then link those containers so that contained EVENTs can be linked by inference.

Phase 2 was similar in complexity to the TempEval-3 Task C challenge from SemEval-2013 in that the classifiers were required to identify relations between events/time expressions \textit{and} to label those relations. Systems only received credit for a narrative container relation if they
found both events/times and correctly assigned a CONTAINS relation between them.

Recall the diversity of classification results observed in Figures~\ref{fig:cleartk2} and  \ref{fig:navytime1}. For consistency, both TempEval-2013 and TempEval-2016 use an evaluation tool, which allows the comparison of results from participating classifiers by using the same metrics of Precision (P), Recall (R), and a harmonic mean thereof (F1 score).
To formalise these following \cite{UzZaman2012} and \cite{bethard2016semeval},
consider the set $S$ of items predicted by the system and the set $H$ of items annotated by the humans.
$S_c$ and $H_c$ are the closed sets of temporal relations, i.e., additional relations that are inferred from other relations.
$S_r$ and $H_r$ are the reduced sets of relations, i.e., with redundant relations removed.\footnote{In Clinical TempEval 2016, sets were not reduced, so $S_r = S$ and $H_r = H$.}
Then, P, R, and F1 score are defined as:

\begin{tabular}{lll}
$P = \cfrac{\mid S_r \cup H_c \mid}{\mid S_r \mid}$ & $R = \cfrac{\mid S_c \cup H_r \mid}{\mid H_r \mid}$ & $F1 = \cfrac{2 \times P \times R}{P + R}$
\end{tabular}
\newline

The $F_1$ score, in particular, is regarded as useful to measure the balance between Precision (which measures how many selected items are relevant) and Recall (which measures how many relevant items are selected). It gives greater importance to low values compared to a simple average. We would ideally like to optimise both Precision and Recall, or at least understand their mutual trade off. \cite{Powers} examines the relationships between these measures, and notes that none takes the number of True Negatives into account. While the nature of the underlying interval algebra in our work, which includes relations and their converses, partly compensates for the focus on P, R and $F_1$ scores, \cite{flach2003geometry} suggests the use of Receiver
Operating Characteristics (ROC) analysis (a concept from Signal Processing) to give a geometric representation of the trade-off between the precision and recall.




\subsection{Inference via Integer Programming}\label{S.IP}

As discussed above, computing transitive closure in Allen's interval algebra is NP-Hard \citep{nebel1995reasoning}.
SemEval participants employ multiple methods to address this complexity,
but generally consider heuristics, which run in time polynomial in the dimension of the input.
In contrast, the focus of our ensemble is to reconcile the reltype labelling predictions of the SemEval participant classifiers,
considering the actual transitive closure.
For that task, we chose Integer Programming (IP), also known as Integer Linear Programming (ILP).
IP is an optimisation technique, with a long history of reach, as documented in \citep{schrijver2003combinatorial,NemhauserWolsey:1985,papadimitriou1998combinatorial},
and there are many well-performing IP solvers.
In particular, it is a leading technique for NP-Hard combinatorial optimisation problems,
where we search for an optimum object within a finite collection.

Usually, IP solvers employ the so called Linear Programming (LP) relaxation with branch-bound-and-cut methods.
In the LP relaxation, a linear function is optimised over an intersection of $m$ linear inequalities and
equalities (so called polyhedron) to find the optimal value for an $n$-vector of decision variables $x_j, j = 1,2,\ldots,n$, i.e.,
\begin{align}
\text{max} & \sum_{j =1}^n\space c_j x_j & \\
\text{s.t. } &
\sum_{j =1}^n a_{i,j} x_j  \leq b_i \hspace{1cm} &  \quad i =1,2,\dots,m\\
&  x_j \in \mathbb{R}^+, & \quad j = 1,2,\ldots,n,
\end{align}
where $c_i, b_i$ and $a_{i,j}$ are known real-valued coefficients and $\mathbb{R}^+$ denotes non-negative real numbers. 
Whereas this continuous relaxation is readily solvable in time polynomial in the dimensions ($n, m$), the constraint on the variables to be integer render the problem NP-Hard.
Despite the fact that, unless $P = NP$,
there is no polynomial in the number of variables ($n$) and the number of constraints ($m$) that would bound the run-time
of an IP solver in the worst case,
the computational performance of modern IP solvers often turns out to be very good.

IP has hence been applied to a wide range of NP-Hard problems in operations research, software verification, mathematical logic, combinatorics, and recently, computational linguistics. Examples of the latter include text summarisation \citep{Woodsend}, semantic role labelling \citep{Roth2004,Punyakanok1346}, global reconciliation on 
temporal labels \citep{Chambers698}, coupling of local event-event and event-time classifiers 
\citep{Do2012677}.
In these cases, the IP formulation makes it possible to enforce global constraints over subgraphs of intervals only; for example \cite{UzZaman2012} enforces global constraints over three intervals. In our study, the rules of interval algebra are applied over the \emph{whole} test text passage.

In particular, we wish to decide the \quotes{best} arc labelling which is consistent with the given interval information for the complete passage of text by reconciling the recommendations from different classifiers. The decisions are integer in nature, \ie we must select exactly one of the reltypes for arcs as suggested in the transitive closure composition table. 

\section{The Ensemble Method}\label{S.IPensemble}
So far we have described Allen's Interval Algebra, the TimeML temporal information markup language to label temporal events in texts, the use of temporal event graphs to model TimeML annotated text, discussed how NLP classifiers can be evaluated and described decision making by Integer Programming. We now show how the combination of these components support our IP ensemble framework.

In an ensemble method, we use an IP model to combine the results of multiple classifiers in order to improve on their individual performance to exploit the diversity among the classifiers \citep{Florian}. \cite{Dietterich} note that a necessary and sufficient condition for an ensemble of classifiers to be more accurate than any individual member, is that the the constituent classifiers exhibit sufficient accuracy and diversity in predictions.

Leading SemEval-2013 classifiers, as evaluated on their precision and recall,  \cite{Chambers2013,Bethard2013,Laokulrat} can detect the patterns that predict the most likely relation between any two events. However, they do not check for global consistency and may produce conflicting predictions. 
We can, however, use an IP ensemble to reconcile the predictions by enforcing the transitivity rules of the underlying relation algebra. The composite relations for path consistency in our ensemble 
are given in tables such as Table~\ref{tab:allenalgebra}. 

In the case of TempEval 2016 Task 12, the subset of temporal relations in the clinical data did not include all inverse relations.
It was hence necessary to introduce some additional relations to cover all possibilities. For example, if classifier1 detects $I_1$ CONTAINS $I_2$ and classifier2 detects $I_2$ CONTAINS $I_1$, we capture the output of classifier2 as using relation CONTAINS\_INV. Should, however, our eventual output be $I_1$ CONTAINS\_INV $I_2$, in order to be understood by the evaluation program, it has to be encoded as $I_2$ CONTAINS $I_1$. We also note that the inverse of OVERLAP is OVERLAP.

We describe the ensemble framework which takes as its input the output from SemEval participant classifiers, converts the participant classifier TimeML outputs to event graphs, feeds the event graphs to our IP model which serves to reconcile conflicts between the participant classifiers and decide a globally consistent interpretation.

\subsection{The Pipeline}\label{S.Pipe}
\begin{figure} [http!]
\begin{center}
 \includegraphics[width=0.75\linewidth]{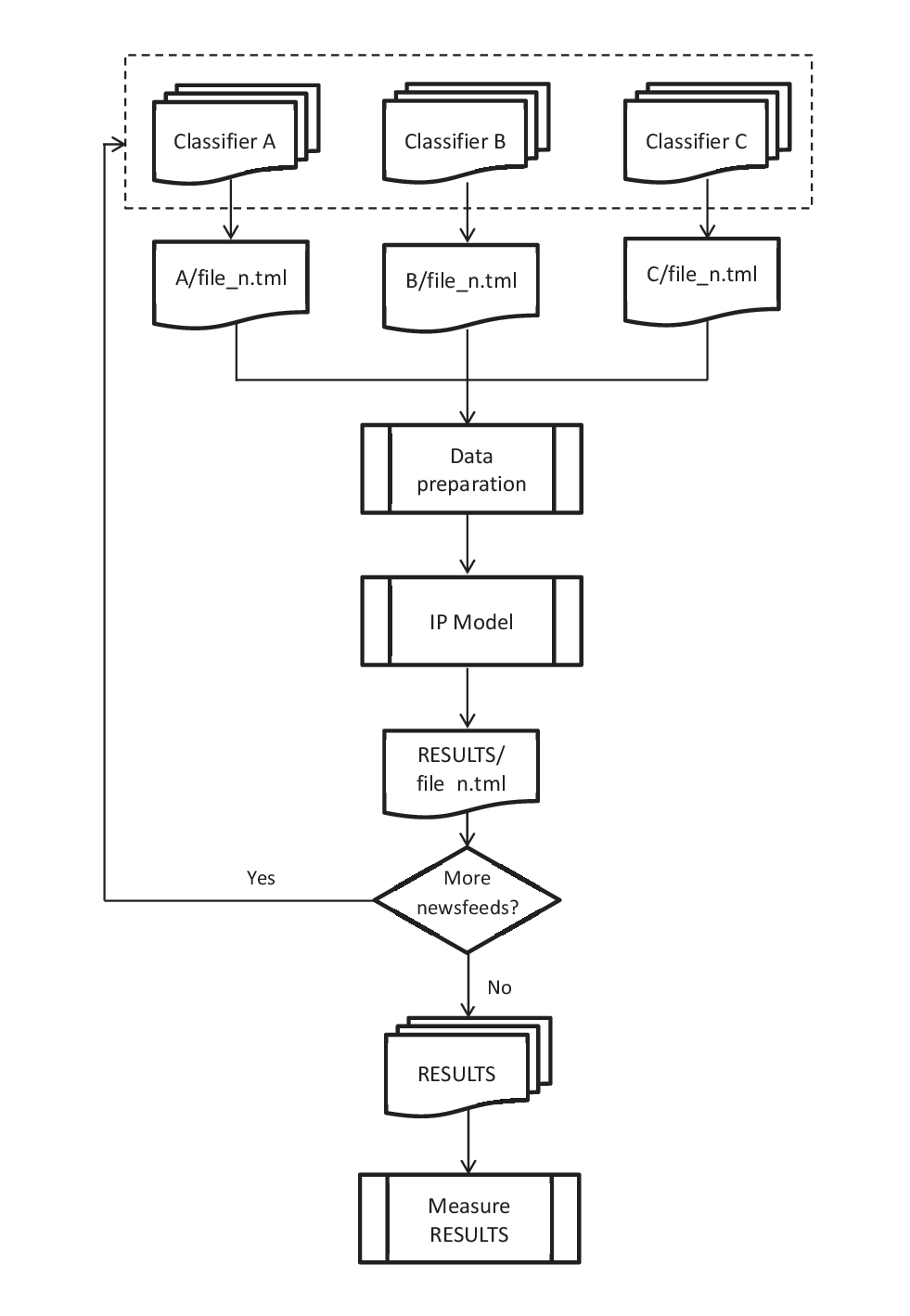}
\caption{A sketch of the ensemble data processing pipeline}
    \label{fig:Pipeline}
    \end{center}
\end{figure}

Figure~\ref{fig:Pipeline} shows an overview of the data flows in our ensemble for a sample of three classifiers $A, B$ and $C$. Variants of ISO TimeML annotation are used in the SemEval competitions. Participants kindly provided their classifications to us. We convert the participant classifiers outputs in TimeML format (which include all the TLINKS) to an event graph (EG) in the data preparation step. We input the EGs to an IP model. We solve the IP model to assign the most likely relation to the EG arcs, subject to global constraints. Finally, we convert the IP solutions to a TimeML-formatted file, which we evaluate with the temporal evaluation tool used in the competition, \emph{temporal\_evaluation.py} from \citep{UzZaman2012Th}. 

\subsection{The Integer Programming Model}\label{S.IP_NLP}
We next describe our IP ensemble formulation.
We use $\mathcal{A}$ to denote the set of annotated TLINKS, \ie the arcs in the transitive closure of an EG, $R$ to denote the reltypes, and ${\mathcal{C}^*}(r_1, r_2): R \times R \to 2^R$ to denote the set of possible composite relationships 
of reltypes $r_1, r_2$. In the case of SemEval-2013 TempEval-3 Task C, $R = \{$p, pi, o, oi, I, Ii, m, mi, s, si, f, fi, NONE$\}$ is the set of reltypes considered, and Table \ref{tab:allenalgebra} captures the corresponding ${\mathcal{C}^*}$.

In the case of TempEval 2016 Task 12, $R = \{$p, pi, c, ci, o, b, bi, NONE$\}$ 
and Table~\ref{tab:intalg} captures the corresponding ${\mathcal{C}^*}$.

Let $x_{i,j}$ be a  binary decision variable:
\begin {equation*}
x_{i,j} =
\begin {cases}
\; 1, \text{ if relationship } j \text{ is assigned to TLINK } i
\\
\; 0, \text{ otherwise}.
\end{cases}
\end{equation*}
The integer program can be seen as a weighted assignment problem
with additional constraints:
\begin{align}
\label{eqn:eqn1}
\text{max} & \sum_{i \in \mathcal{A}}\space \sum_{j \in R}\alpha_{i,j}x_{i,j} & \\
\label{eqn:eqn2}
\text{s.t. } &
\sum_{j \in R}x_{i,j} =1, \hspace{1cm} & \forall \quad i \in \mathcal{A}\\
\label{eqn:eqn3}
&  x_{lm,a} + x_{mn,b} - \sum_{j\in{\mathcal{C}^*}}x_{ln,j} \leq 1,
& \quad  \forall (l,m), (m,n), (l,n) \in{\mathcal{A}}, a, b \in R\\
\label{eqn:eqn4}
& x_{i,j} \in \{0,1\}. 
\end{align}
Eq.~\eqref{eqn:eqn1} is the objective, which maximises the sum of weights assigning TLINK to relation types.


Eq.~\eqref{eqn:eqn2} are mutual exclusivity constraints, which guarantee that exactly one relation type is assigned to each TLINK.
Eq.~\eqref{eqn:eqn3} performs a transitive composition consistency check on each triplet of events $\{l,m,n\}$, where TLINK $(l,m)$ links events $l$ and $m$, $(m,n)$ links events $m$ and $n$, and $(l,n)$ links events $l$ and $n$. Finally, Eq.~\ref{eqn:eqn4} are the binary integrality constraints.

Consider, for example, the arc labelling task in Subfigure~\ref{fig:AllenEx}.c. There is a triplet of events $\{I_1,I_2,I_3\}$ with TLINKs $(I_1,I_2)$ and $(I_2,I_3)$ labelled Contains (c) and Precedes (p), respectively. Eq.~\ref{eqn:eqn2} says we must assign a RELTYPE to $(I_1,I_3)$:

\begin{equation*}
x_{I_1I_3,p} + x_{I_1I_3,pi} + x_{I_1I_3,c} + x_{I_1I_3,ci} + x_{I_1I_3,o} + x_{I_1I_3,b} + x_{I_1I_3,bi} + x_{ln,NONE} = 1
\end{equation*}

Eq.~\ref{eqn:eqn3} restricts RELTYPES for arc $(I_1,I_3)$ to the transitive closure of RELTYPES of 
$(I_1,I_2)$ and $(I_2,I_3)$, as per Table~\ref{tab:intalg}:

\begin {equation*}
x_{I_1I_2,c} + x_{I_2I_3,p} - x_{I_1I_3,p} - x_{I_1I_3,c} - x_{I_1I_3,o} - x_{I_1I_3,bi} \leq 1
\end{equation*}

Since the decision variables are binary integers, when $x_{I_1I_2,c} = x_{I_2I_3,p} = 1$, Eq.~\ref{eqn:eqn3} forces the choice from the 4 consistent RELTYPES for $(I_1,I_3)$ in any feasible solution.
The IP solver then finds the feasible solution in our approach. 



Notice that the weights for $\alpha_{i,j}$ are calculated in the data preparation step, which allows for a wealth of non-linear functions to be explored. We note that there are many weighting functions which could be used for the objective function coefficients, and that there are possible combinations of the individual classifiers. Throughout this paper $\alpha_{i,j}$ is the sum of F1 scores of classifiers suggesting the reltype. Other options are to use probability values based on the empirical probability distribution of assignments from the individual classifiers \citep{Chambers698}, but the F1 metric seems more robust, as it considers both precision and recall.



\section{Computational Experiments }\label{S.Results}

We conducted two sets of computational experiments. In the first set of experiments we create IP ensembles for results from the SemEval-2013 TempEval-3 challenge \citep{UzZaman2012}. Results from the SemEval-2013 TempEval-3 Task C look promising,
 but the amount of data is limited to only 20 newsfeeds, split into training/test sets of 10 newsfeeds each. 
To validate the results, we conducted a second set of experiments on the larger datasets from the Clinical Tempeval 2016 Task 12 challenge \citep{bethard2016semeval}.


The experiments were conducted using a pipeline presented in Figure~\ref{fig:Pipeline}. The key building blocks are a code to formulate IP ensembles (Equations \ref{eqn:eqn1}--\ref{eqn:eqn4}), an IP solver, a variety of parsers, and data-conversion routines.
We formulate each IP instance for the ensemble of classifiers in Pyomo \citep{hart2012pyomo} using F1 scores for weights.
 This allows us to switch between IP solvers seamlessly. It also allows for the execution on a variety of platforms, ranging from laptops to clusters.
As an IP solver, we use both cbc, the open-source IP solver \citep{forrest2005cbc}, and IBM ILOG CPLEX 12.4, the state-of-the-art commercial solver, and compare their performance
on a single core of a PC equipped with
an Intel Xeon E7458 processor clocked at 2.4GHz.

\subsection{Results on SemEval-2013 TempEval-3 Task C}\label{S.Results_2013}

Extensive tests were carried out on the SemEval-2013 TempEval-3 Task C dataset, initially using human intuition, with some trial and error, to compose ensembles that  improved upon previous best F1 scores, and subsequently using an automated approach.
We denote the competition participants' classifiers as ``individual classifiers'' in the results below,  to distinguish them from our ensemble classifiers.

The SemEval-2013 TempEval-3 Task C dataset was split into two parts.
The individual classifiers were trained on the training set,
 and evaluated on the Platinum set.
The participants kindly provided their classifications for the Platinum dataset to us.
Table~\ref{tab:sample2} illustrates how some of the leading individual classifiers, our  ensemble, 
and the human annotators (Platinum) annotated the events in the Obama example in Section~\ref{S.TimeML}. 

\begin{table}[http!]
  \centering
  \caption{Sample of classifier TLINKs}
    \begin{tabular}{llll}
    \toprule
          & E2 (scoring) & E4 (leaving) & E2 (scoring) \\
          & E4 (leaving) & T1 (Friday)  & T1 (Friday) \\
    \midrule
    Cleartk-1 & BEFORE & NONE  & NONE \\
    Cleartk-2 & NONE  & NONE  & NONE \\
    Cleartk-3 & BEFORE & NONE  & NONE \\
    Cleartk-4 & BEFORE & NONE  & NONE \\
    Navytime-1 & BEFORE & NONE  & IS\_INCLUDED \\
    Navytime-2 & SIMULTANEOUS & IS\_INCLUDED & NONE \\
    UTTime-4 & SIMULTANEOUS & IS\_INCLUDED & IS\_INCLUDED \\
    UTTime-5 & SIMULTANEOUS & NONE  & IS\_INCLUDED \\
    Ensemble & BEFORE & IS\_INCLUDED & BEFORE \\
    Platinum & BEFORE & NONE  & BEFORE \\
    \bottomrule
    \end{tabular}%
  \label{tab:sample2}%
\end{table}%


It is customary to report the F1 score on the whole of the Platinum set,
 as used in the challenge.
Table~\ref{tab:used} hence summarises the individual classifiers' precision,
recall, and F1 score on the Platinum dataset.

We explore two training and testing procedures. Firstly, the individual classifiers were trained on the training set
and tested on the Platinum set to obtain their F1 scores. However, as this approach is not quite fair, so we also evaluate splitting the Platinum dataset for testing.
Table~\ref{tab:divresults} shows sample results for the first procedure. We summarise the \emph{training} F1 score, precision, and recall
of several variant ensembles on the Platinum set of 20 newsfeeds. Clearly, an ensemble
(C2, C4, U4+5, N1+2)
using 6 of the 11 individual classifiers
(ClearTK-2, ClearTK-4, UTTime-4, UTTime-5, and Navytime-1 and Navytime-2),
results in the best performance with F1 of 0.3899 and recall of 0.5.
These \emph{training} scores are considerably better than the \emph{test} performance of the
individual classifiers, as summarized in Table~\ref{tab:used}, but we do note that this comparison is not quite fair.

In the second of the two procedures, we split the Platinum dataset into two subsets, S1 and S2,
of ten newsfeeds each.
The individual classifiers, which were trained on the original training set, were evaluated on S1, as detailed in Table~\ref{tab:S2used}.
Table~\ref{tab:S2results} and \ref{tab:S2resultsAutomated} then detail the \emph{test} F1 scores, precision, and recall,
as tested on the 10 newsfeeds of S2.
Throughout, the \emph{test} scores on S2 are considerably better than the \emph{test} performance of the
individual classifiers on $S2$, as presented in Table~\ref{tab:S2used}.
Notice that the F1 score of each individual classifier is worse on S2 than on $S1 \cup S2$, 
as per Tables~\ref{tab:used}--\ref{tab:S2used}, which may suggest why
results of the ensembles are often worse on S2 than on $S1 \cup S2$, as per
Tables~\ref{tab:divresults}--\ref{tab:S2resultsAutomated}.
Overall, the results of the ensembles on $S2$ in Tables~\ref{tab:S2results} and \ref{tab:S2resultsAutomated}
provide convincing evidence of the benefit of the ensemble method.

In particular, Table~\ref{tab:S2results} captures an effort to construct the ensemble iteratively, with a human in the loop.
Considering that good ensembles are composed of classifiers from each of the classifier groups,
 the effort started with ClearTK-2, UTTime-4, and Navytime-1.
(We use the identifiers C2, U4, and N1 introduced in Table~\ref{tab:used}.)
Subsequently, the effort added further classifiers, although the marginal improvement varied.
Notice, for example, that the ensemble C2-U4-N1, which is diverse, outperforms C1-C2-C3-C4, where the classifiers provide similar results, despite the higher F1 scores of the individual classifiers.
Throughout the effor documented in Table~\ref{tab:S2results}, we have used the F1 score of the individual classifiers on S1, the first subset, as weights
in the ensembles, which 
were tested on S2, the second subset.

Next, Table~\ref{tab:S2resultsAutomated} documents an automated search for the best ensemble.
There, we have sampled the performance of 6105 ensembles,
ranging from 2 to 11 individual classifiers, with weight being a convex combination of
precision and recall.
In particular, coefficient of 1.0 in the weight column in Table~\ref{tab:S2resultsAutomated}
suggests weighing with precision of the individual classifiers on S1, the first subset; 0.0
  suggests weighing with recall on S1; and 0.5 suggests weighing with their arithmetic mean on S1.
The precision, recall, and F1 scores reported in Table~\ref{tab:S2resultsAutomated} were tested on S2,
the second subset.
The F1 scores exceeding 0.39 improve upon the human-in-the-loop effort
and the individual classifiers, considerably.\footnote{
We note these numbers are directly comparable to the performance of individual classifiers on S2,
as captured in Tables~\ref{tab:S2used} and \ref{tab:S2results}, but not the performance of the individual classifiers on $S1 \cup S2$,
as documented in Table~\ref{tab:used}.}
One may notice that good ensembles can be obtained with each of the weights tested:
for example the classifier C1-C2-U4-U5 delivers the same F1 score of 0.3903, independent of whether one uses precision, recall, or their mean as the weights.
We hence ascribe the improvement largely to the consistency constraints.

\begin{figure}
	\centering
\caption{SemEval-2013 TempEval-3 Task C results: Results of the second procedure for a sample of ensembles,
with each ensemble's results represented by one data point in the precision-recall plane.
In black or blue, we plot the results of ensembles, which were not dominated by other ensembles in the Pareto sense.
In blue, in particular, we plot the ensembles of Table~\ref{tab:S2resultsAutomated}.
In green, we plot the results of ensembles of Table~\ref{tab:S2results} for comparison.
In red, we plot the results of the individual classifiers,
described by IDs introduced in Table~\ref{tab:S2used}, for comparison.
}
\includegraphics[page=5,width=\textwidth]{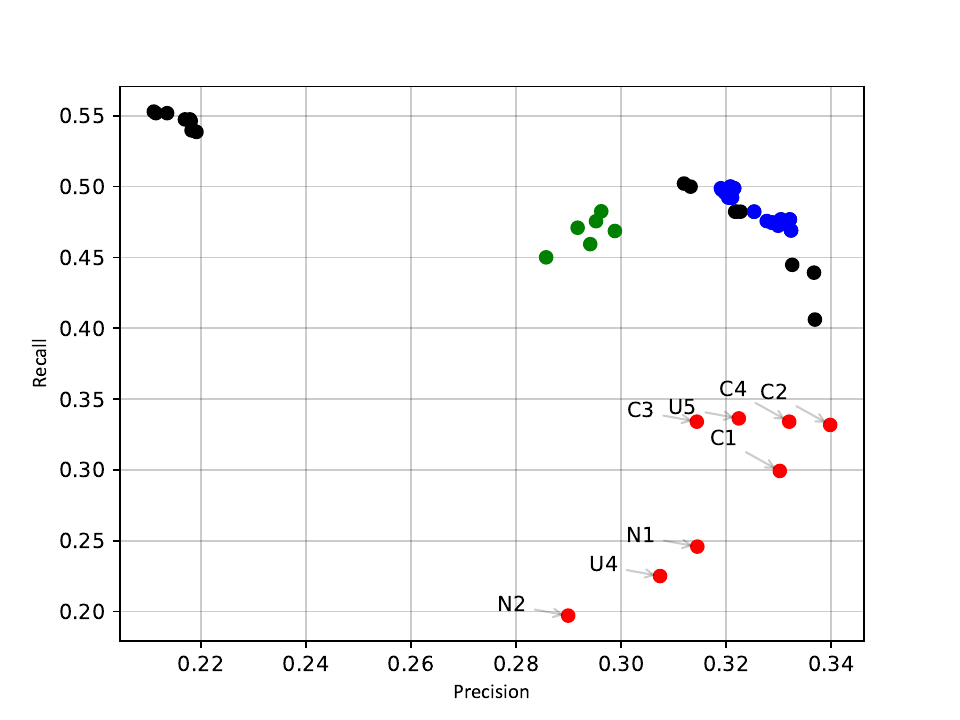}
\label{fig:roc1}
\end{figure}
\begin{figure}
	\centering
\caption{SemEval-2013 TempEval-3 Task C results: Results of the second procedure for a sample of ensembles,
with each ensemble represented by one data point in the plot.
In black, we plot results of ensembles, whose scores were not dominated by other ensembles.
In gray and partially transparent, we plot results of other ensembles.}
\includegraphics[page=2,width=\textwidth]{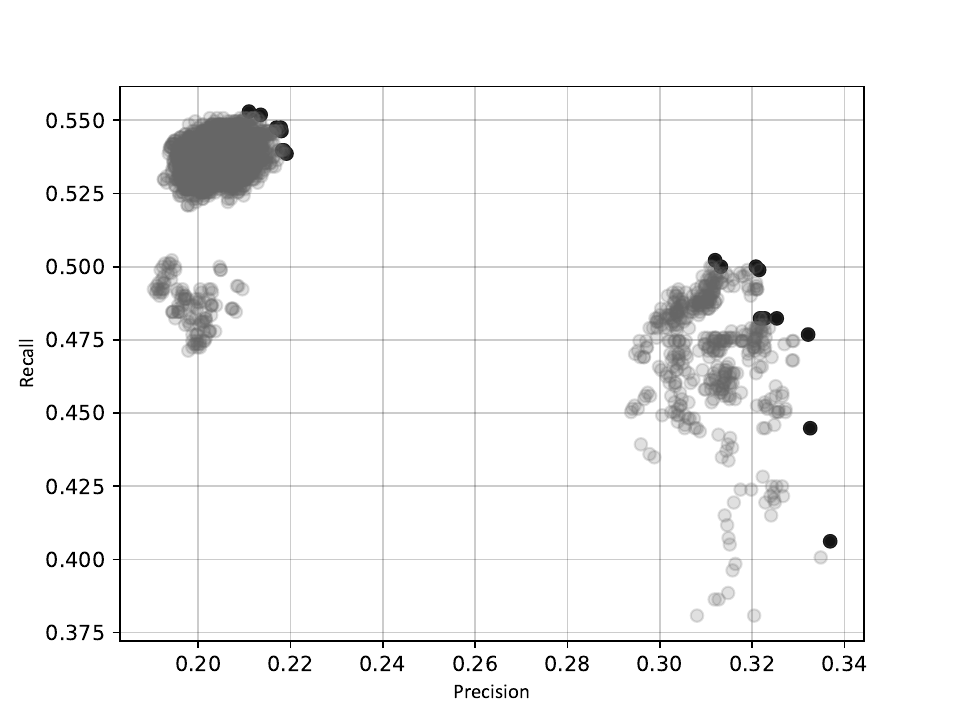}
\label{fig:roc2}
\end{figure}

The automated search yields a number of further insights into the structure of the ``search space'',
 as illustrated in Figures~\ref{fig:roc1} and \ref{fig:roc2}.
In Figure~\ref{fig:roc1}, we plot results of ensembles,
whose scores were not dominated by other ensembles in the sample (in the Pareto sense) in black or blue.
In blue, in particular, we plot the ensembles of Table~\ref{tab:S2resultsAutomated}.
For comparison, we plot the results of ensembles of Table~\ref{tab:S2results} in green
and the results of the individual classifiers in Table~\ref{tab:S2results} in red.
This suggests the shape of the receiver operating characteristic (ROC) curve\footnote{
The ROC curve is usually plotted in terms of recall (true positive rate) and the false positive rate, rather than precision.
We use recall and precision to keep the values directly comparable with Tables~\ref{tab:used}--\ref{tab:S2resultsAutomated}.
}.
As should be expected from classifier with a small, finite number of classes, this is not a proper curve,
but rather a finite collection of points;
indeed, even if you vary the weights continuously, you can obtain only a finite number of outcomes.
Further, the points are clustered, based on the composition of the constituent individual classifiers.
In Figure~\ref{fig:roc2}, we also include the results of dominated ensembles
(in gray and partially transparent), where the clustering is yet more pronounced.


\begin{table} [http!]
  \centering
  \caption{SemEval-2013 TempEval-3 Task C results: The performance of the individual classifiers on the Platinum dataset.}
    \begin{tabular}{llccc}
    \toprule
    ID & Classifier & F1    & Precision & Recall \\
    \midrule
    C1 & cleartk-1 & 0.3517 & 0.3764 & 0.3300 \\
    C2 & cleartk-2 & 0.3624 & 0.3732 & 0.3521 \\
    C3 & cleartk-3 & 0.3421 & 0.3336 & 0.3510 \\
    C4 & cleartk-4 & 0.3594 & 0.3526 & 0.3664 \\
    N1 & navytime-1 & 0.3079 & 0.3519 & 0.2737 \\ 
    N2 & navytime-2 & 0.3588 & 0.5078 & 0.2774 \\ 
    U1 & UT-1 & 0.2428 & 0.1490 & 0.6556 \\ 
    U2 & UT-2 & 0.2415 & 0.1487 & 0.6424 \\
    U3 & UT-3 & 0.2422 & 0.1507 & 0.6170 \\
    U4 & UT-4 & 0.2882 & 0.3752 & 0.2340 \\ 
    U5 & UT-5 & 0.3499 & 0.3605 & 0.3400 \\ 
    \bottomrule
    \end{tabular}%
  \label{tab:used}%
\end{table}%

\begin{table}[http!]
  \centering
  \caption{SemEval-2013 TempEval-3 Task C results: The performance of the individual classifiers on S2, the subset of Platinum.}
    \begin{tabular}{llccc}
    \toprule
    ID & Classifier & F1    & Precision & Recall \\
    \midrule
    C1 & cleartk-1 & 0.3140 & 0.3302 & 0.2993 \\
    C2 & cleartk-2 & 0.3358 & 0.3398 & 0.3318 \\
    C3 & cleartk-3 & 0.3240 & 0.3144 & 0.3341 \\
    C4 & cleartk-4 & 0.3331 & 0.3320 & 0.3341 \\
    N1 & navytime-1 & 0.2760 & 0.3145 & 0.2459 \\ 
    N2 & navytime-2 & 0.2347 & 0.2899 & 0.1972 \\ 
    U4 & UT-4 & 0.2599 & 0.3074 & 0.2251 \\ 
    U5 & UT-5 & 0.3293 & 0.3224 & 0.3364 \\ 
    \bottomrule
    \end{tabular}%
  \label{tab:S2used}%
\end{table}%

\begin{table}[http!]
  \centering
  \caption{SemEval-2013 TempEval-3 Task C results: Results of the first procedure with ensembles described by IDs introduced in
Table~\ref{tab:used}.}
    \begin{tabular}{lccc}
    \toprule
    IDs & F1    & Precision & Recall \\
    \midrule
	C1--3           & 0.3583 & 0.3515 & 0.3653 \\
	C1--4	        & 0.3600 & 0.3488 &	0.3720 \\
	C2, C4, U4, N1	        & 0.3756 & 0.3126 &	0.4702 \\
	C2, U4, N1        & 0.3786 & 0.3205 &	0.4625 \\
	C2, C4, U4+5, N1       & 0.3862 & 0.3182 &	0.4912 \\
	C1--4, U4+5, N1+2	& 0.3877 & 0.3166 &	0.5000 \\
	C1--2, C4, U4+5, N1+2	& 0.3893 & 0.3192 &	0.4989 \\
	C2, C4, U4+5, N1+2	    & 0.3899 & 0.3195 &	0.5000 \\
    \bottomrule
    \end{tabular}%
  \label{tab:divresults}%
\end{table}%

\begin{table}[http!]
  \centering
  \caption{SemEval-2013 TempEval-3 Task C results: The results of the second procedure with manually-picked ensembles described by IDs introduced in Table~\ref{tab:S2used}.}
    \begin{tabular}{lccc}
    \toprule
    IDs & F1    & Precision & Recall \\
    \midrule
	C2, U4, N1	        & 0.3495 & 0.2857 &	0.4501 \\
	C2+4, U4, N1        & 0.3586 & 0.2941 &	0.4594 \\
	C2+4, U4+5, N1+2    & 0.3602 & 0.2917 &	0.4710 \\
	C1--4, U4+5, N1+2	& 0.3643 & 0.2952 &	0.4756 \\
	C2+4, U4+5, N1   	& 0.3649 & 0.2988 &	0.4687 \\
	C1+2, C4, U4+5, N1+2 & 0.3671 & 0.2962 & 0.4826 \\
    \bottomrule
    \end{tabular}%
  \label{tab:S2results}%
\end{table}%

\begin{table}[http!]
  \centering
  \caption{SemEval-2013 TempEval-3 Task C results: A selection of results of the second procedure with automatically constructed ensembles, described by IDs introduced in Table~\ref{tab:S2used}, with weight 1.0 suggesting weighing with precision, weight 0.0
  suggesting weighing with recall, and 0.5 weighing with their arithmetic mean.}
    \begin{tabular}{llccc}
    \toprule
    IDs & weight & F1    & Precision & Recall \\
    \midrule
C1-C2-U4-U5 & 0.0 & 0.3881 & 0.3277 & 0.4757 \\
C1-N1-U5 & 0.5 & 0.3882 & 0.3204 & 0.4923 \\
C1-N1-U5 & 1.0 & 0.3882 & 0.3204 & 0.4923 \\
C2-U5 & 1.0 & 0.3884 & 0.3287 & 0.4746 \\
C2-U5 & 0.5 & 0.3884 & 0.3287 & 0.4746 \\
C1-C2-U5 & 1.0 & 0.3884 & 0.3287 & 0.4746 \\
C1-C2-U5 & 0.5 & 0.3884 & 0.3287 & 0.4746 \\
C2-U5 & 0.0 & 0.3885 & 0.3299 & 0.4724 \\
C1-N1-U5-N2 & 0.0 & 0.3885 & 0.3209 & 0.4923 \\
C2-U4-U5 & 0.0 & 0.3886 & 0.3289 & 0.4746 \\
C2-C4-U4-U5 & 1.0 & 0.3886 & 0.3253 & 0.4823 \\
C1-N1-U5 & 0.0 & 0.3886 & 0.3211 & 0.4923 \\
C1-N1-U4-U5-N2 & 0.0 & 0.3887 & 0.3211 & 0.4923 \\
C2-N1-U5 & 0.0 & 0.3887 & 0.3198 & 0.4956 \\
C1-N1-U5-N2 & 1.0 & 0.3889 & 0.3209 & 0.4934 \\
C1-N1-U5-N2 & 0.5 & 0.3889 & 0.3209 & 0.4934 \\
C1-C2-N1-U4-U5 & 0.0 & 0.3889 & 0.3191 & 0.4978 \\
C1-U5 & 0.5 & 0.3890 & 0.3323 & 0.4691 \\
C1-U5 & 1.0 & 0.3890 & 0.3323 & 0.4691 \\
C2-N1-U5-N2 & 0.0 & 0.3891 & 0.3199 & 0.4967 \\
C1-C2-N1-U4-U5-N2 & 0.5 & 0.3892 & 0.3190 & 0.4989 \\
C1-C2-N1-U4-U5-N2 & 1.0 & 0.3892 & 0.3190 & 0.4989 \\
C1-N1-U4-U5-N2 & 1.0 & 0.3892 & 0.3209 & 0.4945 \\
C1-N1-U4-U5-N2 & 0.5 & 0.3892 & 0.3209 & 0.4945 \\
C1-C2-N1-U5 & 0.0 & 0.3903 & 0.3210 & 0.4978 \\
C1-C2-U4-U5 & 1.0 & 0.3903 & 0.3304 & 0.4768 \\
C1-C2-U4-U5 & 0.5 & 0.3903 & 0.3304 & 0.4768 \\
C1-C2-N1-U4-U5-N2 & 0.0 & 0.3908 & 0.3208 & 0.5000 \\
C1-C2-N1-U5-N2 & 0.0 & 0.3910 & 0.3215 & 0.4989 \\
C1-C2-U5 & 0.0 & 0.3915 & 0.3321 & 0.4768 \\
    \bottomrule
    \end{tabular}%
  \label{tab:S2resultsAutomated}%
\end{table}%

\subsection{Results on TempEval 2016 Task 12}\label{S.Semeval2016}

We obtained output data from the seven leading participants in this task, including 13 different classifiers\footnote{UtahBMI submitted corrected classifiers to the Task 12 challenge, whose results were too late for formal inclusion. We opted to use this data instead.}.
Overall, the clinical data was composed of 151 files, 100 of which were used in the narrative container relations part of the challenge. We split the data from each participating classifier and from the ``gold'', human-annotated data into ten randomly sampled $A/B$ datasets of 50 files each, where $A_n = \overline{\rm B_n}$ and $n \in {0..9}$. The dataset $A_n$ from each classifier (and from the ``gold'' set) contained the same set of files; likewise for each dataset $B_n$. Each classifier dataset $A_n$ (or $B_n$) was evaluated against the corresponding ``gold'' $A_n$ (or $B_n$) set using the same evaluation program that was used in SemEval-2016. The F1 score of each $A_n$ set was used as the weight in ensembles built from the corresponding $B_n$ set, and vice versa. Testing of ensembles was carried out against all twenty datasets and results were compared against the corresponding twenty ``gold'' sets.
The classifier uthealth-p2s1 \citep{lee2016uthealth} achieved the highest F1 score in the Task 12 challenge, with an F1 score of 0.573. This classifier also scored the highest in each of the twenty randomly sampled A/B datasets. We used the F1 scores of each of the twenty datasets for this classifer as the target to improve upon with our ensemble.

Table~\ref{tab:targetscores} illustrates the scores of the individual classifiers against the hand-annotated set for the full set and also the average scores for the ten A sets and ten B sets. We used the same evaluation tool \emph{anafora.evaluation.py} that was used in the Task 12 challenge. The best-known scores are highlighted in bold.

\begin{table}[http!]
  \centering
  \caption{SemEval2016 results: Classifier and average A/B sample results.}
    \begin{tabular}{llccccc}
    \toprule
    ID & Classifier &  Precision & Recall & F1 & F1 & F1 \\
     &  &  Overall & Overall & Overall & A set & B set \\
    \midrule
  CC & CDE-IIITH-crf & 0.493 & 0.185 & 0.269 & 0.271 & 0.268 \\
  CD & CDE-IIITH-deepnl & 0.348 & 0.284 & 0.313 & 0.316 & 0.31 \\
  G & GUIR-Phase2-Run1 & 0.546 & 0.471 & 0.506 & 0.507 & 0.504 \\
  K1 & KULeuven-LIIR-run1 & 0.714 & 0.428 & 0.536 & 0.538 & 0.534 \\
  K2 & KULeuven-LIIR-run2 & 0.715 & 0.429 & 0.536 & 0.538 & 0.534 \\
  L1 & LIMSI\_COT-RUN1 & 0.704 & 0.436 & 0.538 & 0.54 & 0.537 \\
  L2 & LIMSI\_COT-RUN2 & \textbf{0.751} & 0.32 & 0.449 & 0.455 & 0.444 \\
  Ub1 & UtahBMI-RUN1-corrected & 0.711 & 0.372 & 0.489 & 0.489 & 0.489 \\
  Ub2 & UtahBMI-RUN2-corrected & 0.693 & 0.425 & 0.527 & 0.528 & 0.525 \\
  Ut1 & uthealth-p2s1 & 0.588 & \textbf{0.559} & \textbf{0.573} & 0.573 & 0.572 \\
  Ut2 & uthealth-p2s2 & 0.568 & 0.564 & 0.566 & 0.566 & 0.565 \\
  V1 & VUACLTL-run1-phase2 & 0.642 & 0.345 & 0.449 & 0.451 & 0.448 \\
  V2 & VUACLTL-run2-phase2 & 0.589 & 0.368 & 0.453 & 0.454 & 0.452 \\
    \bottomrule
    \end{tabular}%
  \label{tab:targetscores}%
\end{table}%

\subsubsection{Construction of the Semeval2016 Ensemble}\label{S.Semeval2016results}
Our ensembles were built in a piece-wise fashion, with measurements taken at each stage to determine whether the results were an improvement on previous ensembles. As a starting point, we selected the three highest-scoring classifiers, from three different participants
uthealth-p2s1, LIMSI\_COT-RUN1 \citep{grouin2016limsi} and KULeuven-LIIR-run1 \citep{leeuwenberg2016kuleuven}.
Subsequently, we added classifiers, one at a time. As expected, recall continued to improve, but precision deteriorated, due to the introduction of irrelevant TLINKs. At some point, improved recall is outweighed by poor precision and the F1 score deteriorates. We addressed this issue by filtering the classifiers according to a precision score threshold. Classifiers above the threshold can nominate new relations to the ensemble \textit{and} vote on the label. Classifiers below the threshold can only vote. This approach yielded further improvements in the F1 score of the ensemble.
The best score was obtained using uthealth-p2s1, LIMSI\_COT-RUN1, KULeuven-LIIR-run1, GUIR-Phase2-Run1 \citep{cohan2016guir} and UtahBMI-RUN1-corrected \citep{khalifa2016utahbmi}.
Interestingly, we achieved a good result (0.592) when we used a seven-classifier ensemble, adding CDE-IIITH-crf \citep{chikka2016cde} and VUACLTL-run1-phase2 \citep{caselli2016vuacltl}, once again supporting the assertion that diversity matters.

\begin{figure}
	\centering
\caption{The results of selected ensembles on Semeval2016.
In black, we plot results of ensembles, whose scores were not dominated by other ensembles within the sample.
In gray and partially transparent, we plot results of other ensembles.}
\includegraphics[page=2,width=\textwidth]{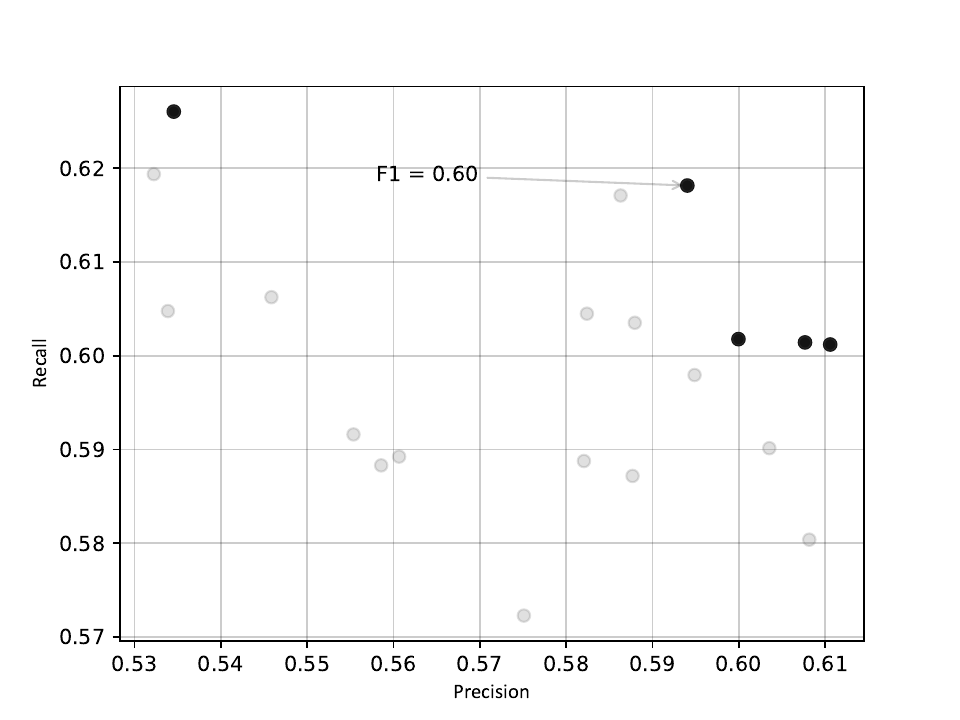}
\label{fig:roc3}
\end{figure}

Let us illustrate the ensemble construction procedure in Table~\ref{tab:clinicalensembleresults}. There results are listed in the order, in which the ensembles were tested. The symbol $\bigtriangleup$ indicates that the F1 score is better than any previous F1 scores. At each stage, the results were checked to see if the ensemble score was improving. If there was a deterioration, we increased the precision threshold, or replaced a classifer with an alternative one. Results indicate that a precision threshold of 0.65 works best. If we increased the threshold to 0.70, the recall score suffered, because too few classifiers were able to nominate relations to the ensemble. Further work might be undertaken to establish the optimum precision threshold, as it obviously depends on the precision of the classifiers available to the ensemble and other factors.
Figure~\ref{fig:roc3} illustrates that this procedure actually performs rather well, in comparison with sampling the individual classifiers randomly.

\begin{table}[th]
  \centering
  \caption{SemEval2016 results: The results of individual ensembles and precision threshold values}
    \begin{tabular}{lccccc}
    \toprule
    Ensemble & Precision  & Precision & Recall & F1 &   \\
    composition & threshold  &  &  & score &   \\
    \midrule
  Uthealth-p2s1 & 0.00 & 0.588 & 0.559 & 0.573 &   \\
  Ut1/L1/K1 & 0.00 & 0.547 & 0.633 & 0.587 & $\bigtriangleup$  \\
  Ut1/L1/K1/Ub2 & 0.00 & 0.532 & 0.646 & 0.583 &   \\
  Ut1/L1/K1/Ub2 & 0.60 & 0.575 & 0.606 & 0.588 & $\bigtriangleup$  \\
  Ut1/L1/K1/Ub1 & 0.60 & 0.586 & 0.597 & 0.589 & $\bigtriangleup$  \\
  Ut1/L1/G/K1/Ub1 & 0.60 & 0.587 & 0.599 & 0.590 & $\bigtriangleup$  \\
  Ut1/L1/G/K1/V1/Ub1 & 0.60 & 0.553 & 0.613 & 0.579 &   \\
  Ut1/L1/G/K1/V1/Ub1 & 0.65 & 0.605 & 0.580 & 0.592 & $\bigtriangleup$  \\
  Ut1/L1/G/K1/V1/Ub1 & 0.70 & 0.636 & 0.533 & 0.575 &   \\
  Ut1/L1/K1/V1/Ub1 & 0.65 & 0.605 & 0.578 & 0.591 &   \\
  Ut1/L1/CC/G/K1/V1/Ub1 & 0.65 & 0.605 & 0.581 & 0.592 & =  \\
  Ut1/L1/K1/Ub1 & 0.65 & 0.615 & 0.574 & 0.594 & $\bigtriangleup$  \\
  Ut1/L1/G/K1/Ub1 & 0.65 & 0.616 & 0.576 & 0.595 & $\bigtriangleup$  \\
  \bottomrule
    \end{tabular}%
  \label{tab:clinicalensembleresults}%
\end{table}%

\subsubsection{Analysis of Effect Size}
Table~\ref{tab:bestscore} presents another analysis of the F1 score of the ensemble.
We used Cohen's \textit{d} measure to
evaluate the \textit{effect size}, or the strength of the improvement considering the variability in the data.
Cohen's \textit{d} 
ia the standardised difference between two means:
$$d = \cfrac{|\overline{x_1} - \overline{x_2}|}{s_p}$$
where $s_p$ is the pooled standard deviation of the samples of sizes $n_1, n_2$, respectively, with standard deviations $s_1, s_2$, respectively, is:
$$s_p = \sqrt{\cfrac{(n_1-1)s_1^2 + (n_2-1)s_2^2}{n_1 + n_2 -2}}$$

A Cohen's \emph{d} value greater than $|1.2|$ is described as \quotes{very large}, while greater than $|2.0|$ is described as ``huge'' \citep{sawilowsky2009new}. Table~\ref{tab:bestscore} shows the effect of the ensembles considering either Precision or Recall is very large, while the ensemble considering the F1 score is huge.

\begin{table}[th]
  \centering
  \caption{Semeval2016 results: Results of ensemble composed of classifiers Ut1/L1/G/K1/Ub1.}
    \begin{tabular}{ccccccc}
    \toprule
   Testset & Classifier & Ensemble & Classifier & Ensemble & Classifier & Ensemble  \\
    & Precision & Precision & Recall & Recall & F1 score & F1 score  \\
    \midrule
  00-A & 0.572 & 0.615 & 0.549 & 0.563 & 0.561 & 0.588  \\
  01-A & 0.603 & 0.630 & 0.543 & 0.569 & 0.572 & 0.598  \\
  02-A & 0.604 & 0.636 & 0.570 & 0.578 & 0.586 & 0.606  \\
  03-A & 0.588 & 0.608 & 0.553 & 0.573 & 0.570 & 0.590  \\
  04-A & 0.566 & 0.601 & 0.542 & 0.570 & 0.554 & 0.585  \\
  05-A & 0.597 & 0.636 & 0.560 & 0.581 & 0.578 & 0.607  \\
  06-A & 0.576 & 0.595 & 0.576 & 0.594 & 0.576 & 0.594  \\
  07-A & 0.604 & 0.623 & 0.569 & 0.598 & 0.586 & 0.610  \\
  08-A & 0.573 & 0.619 & 0.572 & 0.595 & 0.573 & 0.607  \\
  09-A & 0.603 & 0.619 & 0.547 & 0.565 & 0.574 & 0.591  \\
  00-B & 0.604 & 0.617 & 0.568 & 0.589 & 0.585 & 0.603  \\
  01-B & 0.569 & 0.598 & 0.579 & 0.585 & 0.574 & 0.592  \\
  02-B & 0.569 & 0.593 & 0.545 & 0.573 & 0.556 & 0.583  \\
  03-B & 0.588 & 0.625 & 0.564 & 0.579 & 0.576 & 0.601  \\
  04-B & 0.606 & 0.630 & 0.573 & 0.581 & 0.589 & 0.604  \\
  05-B & 0.578 & 0.597 & 0.557 & 0.569 & 0.567 & 0.583  \\
  06-B & 0.599 & 0.638 & 0.543 & 0.559 & 0.570 & 0.596  \\
  07-B & 0.571 & 0.608 & 0.547 & 0.552 & 0.559 & 0.579  \\
  08-B & 0.601 & 0.613 & 0.547 & 0.559 & 0.573 & 0.585  \\
  09-B & 0.573 & 0.614 & 0.571 & 0.586 & 0.572 & 0.600  \\
  \midrule
  Mean & 0.587 & 0.616 & 0.559 & 0.576 & 0.573 & 0.595  \\
  Std.\ dev.\ & 0.015 & 0.014 & 0.013 & 0.013 & 0.010 & 0.009  \\
  Cohen's \emph{d}  & &2.00 & &1.31 & & 2.34  \\
    \bottomrule
    \end{tabular}%
  \label{tab:bestscore}%
\end{table}%


\subsection{A Remark on IP Run-time}
Last but not least, let us remark on the run-time of the IP solvers.
Using CPLEX, we were able to solve even the largest IP ensemble instances in seconds. Table~\ref{tab:ipperf} shows the run-time performance of the IP ensemble that reconciles all 11 SemEval-2013 TempEval-3 Task C individual classifiers (see Table~\ref{tab:used}) on each of the 20 newsfeeds in the challenge.
This should not be surprising, considering the largest instance (WSJ\_20130322\_159) for the ensemble of all 11 individual classifiers had dimension of the decision variable as low as 7365. 
The IP ensemble instances  for SemEval-2016 Task 12 (Clinical TempEval) were solved faster still, due to the sparsity of the EG. This suggests the scalability of our IP ensemble approach to related temporal reasoning problems.

\begin{table}[htb!]
  \centering
  \caption{IP ensemble performance.}
    \begin{tabular}{lrrr}
    \toprule
		 & & cbc & CPLEX \\
		Instance & Dim. & (secs)   & (secs) \\
	\midrule
		AP\_20130322 & 2,790 & 8.01   & 0.57 \\
		bbc\_20130322\_1150 & 4,155 & 23.94   & 1.02 \\
		bbc\_20130322\_1353 & 4,440 & 119.1   & 3.73 \\
		bbc\_20130322\_1600 & 2,535 & 16.37   & 1.49 \\
		bbc\_20130322\_332 & 3,675 & 8.56   & 1.01 \\
		bbc\_20130322\_721 & 2,115 & 11.16   & 0.69 \\
		CNN\_20130322\_1003 & 7,200 & 181.49   & 4.69 \\
		CNN\_20130322\_1243 & 900 & 0.9   & 0.14 \\
		CNN\_20130322\_248 & 2,025 & 2   & 0.71 \\
		CNN\_20130322\_314 & 3,045 & 19.86   & 1.19 \\
		CNN\_20130322\_821 & 285 & 0.04   & 0.01 \\
		nyt\_20130321\_cyprus & 6,285 & 82.55   & 4.11 \\
		nyt\_20130321\_china & 4,755 & 55.61   & 3.76 \\
		nyt\_20130321\_sarkozy & 2,130 & 11.16   & 2.43 \\
		nyt\_20130321\_women & 4,950 & 41.63   & 4.04 \\
		nyt\_20130321\_strange & 2,820 & 34.82   & 1.89 \\
		WSJ\_20130318\_731 & 2,025 & 5.76   & 0.43 \\
		WSJ\_20130321\_1145 & 1,965 & 6.58   & 0.64 \\
		WSJ\_20130322\_804 & 3,270 & 81.87   & 2.56 \\
		WSJ\_20130322\_159 & 7,365 & n/a   & 75.57 \\
    \bottomrule
    \end{tabular}%
  \label{tab:ipperf}%
\end{table}%

\section{Conclusions}\label{S.Conclusion}
Building an ensemble of classifiers can provide a significant improvement in precision and recall over the individual classifiers.
The improvement in recall is understandable, considering the ensemble uses the \emph{union} of classifiers' results.

We have demonstrated that enforcing consistency constraints over the results of even a small number of individual classifiers improves precision
 over the individual classifiers in two recent competitions in classifying temporal relations.
This overall consistency is required in order to make practical use of the inferred temporal relations between events in real-world applications.
While one cannot hope for the processing of web-scale texts using the present IP solvers in milliseconds,
the approach seems well-suited to the scale of the data sets used in the competitions, which in turn may be representative of real-world datasets.

Promising directions for future research include exploring alternative weights such as precision, alternative means of combining the individual classifiers, cf. \cite{667881}, and the use of soft constraints, cf. \cite{Burke2201271}.
While this research concentrated on event and temporal relations detection and classification, it can readily be extended to other types of classifiers, such as for spatial expressions, where a similar algebra can be developed in terms of geo-coordinates.
One may also consider replacing integer programming with constraint logic programming (CLP) or valued constraint satisfaction (VCS),  
or a binary search over the objective function value coupled with constraint satisfaction (CSP) and propositional satisfiability (SAT).
There, some of the solvers may also present compelling cost-benefit trade-offs.


\paragraph*{Acknowledgements}
\quad The authors would like to thank \cite{Chambers2013,Bethard2013,Laokulrat,leeuwenberg2016kuleuven,caselli2016vuacltl,chikka2016cde,grouin2016limsi,khalifa2016utahbmi,cohan2016guir}, and
\cite{lee2016uthealth}, whose work and data kindly shared with us made this research possible.
Jakub Marecek has received funding from the European Union Horizon 2020 Programme (Horizon2020/2014-2020), under grant agreement no. 688380.

\bibliographystyle{plainnat} 
\bibliography{refs_NLP}

\begin{thebibliography}{49}
\providecommand{\natexlab}[1]{#1}
\providecommand{\url}[1]{\texttt{#1}}
\expandafter\ifx\csname urlstyle\endcsname\relax
  \providecommand{\doi}[1]{doi: #1}\else
  \providecommand{\doi}{doi: \begingroup \urlstyle{rm}\Url}\fi

\bibitem[Allen(1983)]{Allen}
James~F Allen.
\newblock Maintaining knowledge about temporal intervals.
\newblock \emph{Communications of the {ACM}}, 26\penalty0 (11):\penalty0
  832--843, 1983.

\bibitem[Bethard(2013)]{Bethard2013}
Steven Bethard.
\newblock Cleartk-timeml: A minimalist approach to tempeval 2013.
\newblock In \emph{Second Joint Conference on Lexical and Computational
  Semantics (* SEM)}, volume~2, pages 10--14, 2013.

\bibitem[Bethard et~al.(2016)Bethard, Savova, Chen, Derczynski, Pustejovsky,
  and Verhagen]{bethard2016semeval}
Steven Bethard, Guergana Savova, Wei-Te Chen, Leon Derczynski, James
  Pustejovsky, and Marc Verhagen.
\newblock Semeval-2016 task 12: Clinical tempeval.
\newblock \emph{Proceedings of SemEval}, pages 1052--1062, 2016.

\bibitem[Bhattacharya and Getoor(2007)]{Bhattacharya2007}
Indrajit Bhattacharya and Lise Getoor.
\newblock Collective entity resolution in relational data.
\newblock \emph{ACM Trans. Knowl. Discov. Data}, 1\penalty0 (1), March 2007.
\newblock ISSN 1556-4681.

\bibitem[Bier et~al.(2008)Bier, Card, and Bodnar]{bier2008}
Eric~A Bier, Stuart~K Card, and John~W Bodnar.
\newblock Entity-based collaboration tools for intelligence analysis.
\newblock In \emph{Visual Analytics Science and Technology, 2008. VAST'08. IEEE
  Symposium on}, pages 99--106. IEEE, 2008.

\bibitem[Burke et~al.(2012)Burke, Mare{\v c}ek, Parkes, and Rudov{\'
  a}]{Burke2201271}
Edmund~K Burke, Jakub Mare{\v c}ek, Andrew~J Parkes, and Hana Rudov{\' a}.
\newblock A branch-and-cut procedure for the udine course timetabling problem.
\newblock \emph{Annals of Operations Research}, 194\penalty0 (1):\penalty0
  71--87, 2012.

\bibitem[Caselli and Morante(2016)]{caselli2016vuacltl}
Tommaso Caselli and Roser Morante.
\newblock Vuacltl at semeval 2016 task 12: A crf pipeline to clinical tempeval.
\newblock \emph{Proceedings of SemEval}, pages 1241--1247, 2016.

\bibitem[Chambers(2013)]{Chambers2013}
Nathanael Chambers.
\newblock Navytime: Event and time ordering from raw text.
\newblock In \emph{Second Joint Conference on Lexical and Computational
  Semantics (*SEM), Volume 2}, pages 73--77. Association for Computational
  Linguistics, 2013.

\bibitem[Chambers and Jurafsky(2008)]{Chambers698}
Nathanael Chambers and Daniel Jurafsky.
\newblock Jointly combining implicit constraints improves temporal ordering.
\newblock In \emph{Proceedings of the 2008 Conference on Empirical Methods in
  Natural Language Processing}, pages 698--706. Association for Computational
  Linguistics, 2008.

\bibitem[Chikka(2016)]{chikka2016cde}
Veera~Raghavendra Chikka.
\newblock Cde-iiith at semeval-2016 task 12: Extraction of temporal information
  from clinical documents using machine learning techniques.
\newblock \emph{Proceedings of SemEval}, pages 1237--1240, 2016.

\bibitem[Cohan et~al.(2016)Cohan, Meurer, and Goharian]{cohan2016guir}
Arman Cohan, Kevin Meurer, and Nazli Goharian.
\newblock Guir at semeval-2016 task 12: Temporal information processing for
  clinical narratives.
\newblock \emph{Proceedings of SemEval}, pages 1248--1255, 2016.

\bibitem[Daykin et~al.(2016)Daykin, Miller, and Ryan]{Daykin}
Jacqueline~W. Daykin, Mirka Miller, and Joe Ryan.
\newblock Trends in temporal reasoning: Constraints, graphs and posets.
\newblock In Ilias~S. Kotsireas, Siegfried~M. Rump, and Chee~K. Yap, editors,
  \emph{Mathematical Aspects of Computer and Information Sciences}, pages
  290--304, Cham, 2016. Springer International Publishing.
\newblock ISBN 978-3-319-32859-1.

\bibitem[Dietterich(2000)]{Dietterich}
Thomas~G Dietterich.
\newblock Ensemble methods in machine learning.
\newblock In \emph{Multiple classifier systems}, pages 1--15. Springer, 2000.

\bibitem[Do et~al.(2012)Do, Lu, and Roth]{Do2012677}
Quang~Xuan Do, Wei Lu, and Dan Roth.
\newblock Joint inference for event timeline construction.
\newblock In \emph{Proceedings of the 2012 Joint Conference on Empirical
  Methods in Natural Language Processing and Computational Natural Language
  Learning}, pages 677--687. Association for Computational Linguistics, 2012.

\bibitem[Dong et~al.(2005)Dong, Halevy, and Madhavan]{Dong2005}
Xin Dong, Alon Halevy, and Jayant Madhavan.
\newblock Reference reconciliation in complex information spaces.
\newblock In \emph{Proceedings of the 2005 ACM SIGMOD International Conference
  on Management of Data}, SIGMOD '05, pages 85--96, New York, NY, USA, 2005.
  ACM.
\newblock ISBN 1-59593-060-4.

\bibitem[Dong et~al.(2013)Dong, Berti-Equille, and Srivastava]{dong2013data}
Xin~Luna Dong, Laure Berti-Equille, and Divesh Srivastava.
\newblock Data fusion: resolving conflicts from multiple sources.
\newblock In \emph{Handbook of Data Quality}, pages 293--318. Springer Berlin
  Heidelberg, 2013.

\bibitem[Flach(2003)]{flach2003geometry}
Peter~A Flach.
\newblock The geometry of roc space: understanding machine learning metrics
  through roc isometrics.
\newblock In \emph{Proceedings of the 20th International Conference on Machine
  Learning (ICML-03)}, pages 194--201, 2003.

\bibitem[Florian et~al.(2002)Florian, Cucerzan, Schaefer, and
  Yarowsky]{Florian}
Radu Florian, Silviu Cucerzan, Charles Schaefer, and David Yarowsky.
\newblock Combining classifiers for word sense disambiguation.
\newblock \emph{Natural Language Engineering}, 8:\penalty0 327--341, 12 2002.

\bibitem[Forrest and Lougee-Heimer(2005)]{forrest2005cbc}
John Forrest and Robin Lougee-Heimer.
\newblock Cbc user guide.
\newblock In \emph{Tutorials in Operations Research}, pages 257--277. INFORMS,
  2005.

\bibitem[Glava\v{s} and \v{S}najder(2015)]{Glavas}
Goran Glava\v{s} and Jan \v{S}najder.
\newblock Construction and evaluation of event graphs.
\newblock \emph{Natural Language Engineering}, 21:\penalty0 607--652, 2015.

\bibitem[Grouin and Moriceau(2016)]{grouin2016limsi}
Cyril Grouin and V{\'e}ronique Moriceau.
\newblock Limsi at semeval-2016 task 12: machine-learning and temporal
  information to identify clinical events and time expressions.
\newblock \emph{Proceedings of SemEval}, pages 1225--1230, 2016.

\bibitem[Hart et~al.(2012)Hart, Laird, Watson, and Woodruff]{hart2012pyomo}
William~E Hart, Carl Laird, Jean-Paul Watson, and David~L Woodruff.
\newblock \emph{Pyomo -- {O}ptimization modeling in python}.
\newblock Springer Science \& Business Media, 2012.

\bibitem[Huang and Lu(2016)]{Huang}
Chung-Chi Huang and Zhiyong Lu.
\newblock Community challenges in biomedical text mining over 10 years:
  success, failure and the future.
\newblock \emph{Briefings in Bioinformatics}, 17\penalty0 (1):\penalty0
  132--144, 2016.
\newblock \doi{10.1093/bib/bbv024}.
\newblock URL \url{+ http://dx.doi.org/10.1093/bib/bbv024}.

\bibitem[Khalifa et~al.(2016)Khalifa, Velupillai, and
  Meystre]{khalifa2016utahbmi}
Abdulrahman Khalifa, Sumithra Velupillai, and Stephane Meystre.
\newblock Utahbmi at semeval-2016 task 12: Extracting temporal information from
  clinical text.
\newblock \emph{Proceedings of SemEval}, pages 1256--1262, 2016.

\bibitem[Kittler et~al.(1998)Kittler, Hatef, Duin, and Matas]{667881}
Josef Kittler, Mohamad Hatef, Robert~P.W. Duin, and Jiri Matas.
\newblock On combining classifiers.
\newblock \emph{IEEE Transactions on Pattern Analysis and Machine
  Intelligence}, 20\penalty0 (3):\penalty0 226--239, 1998.

\bibitem[Ladkin(1990)]{Ladkin}
Peter~B Ladkin.
\newblock \emph{Constraint reasoning with intervals: a tutorial, survey and
  bibliography}.
\newblock International Computer Science Institute, 1990.

\bibitem[Laokulrat et~al.(2013)Laokulrat, Miwa, Tsuruoka, and
  Chikayama]{Laokulrat}
Natsuda Laokulrat, Makoto Miwa, Yoshimasa Tsuruoka, and Takashi Chikayama.
\newblock Uttime: Temporal relation classification using deep syntactic
  features.
\newblock In \emph{Second Joint Conference on Lexical and Computational
  Semantics (* SEM)}, volume~2, pages 88--92, 2013.

\bibitem[Lee et~al.(2016)Lee, Zhang, Xu, Moon, Wang, Wu, and
  Xu]{lee2016uthealth}
Hee-Jin Lee, Yaoyun Zhang, Jun Xu, Sungrim Moon, Jingqi Wang, Yonghui Wu, and
  Hua Xu.
\newblock Uthealth at semeval-2016 task 12: an end-to-end system for temporal
  information extraction from clinical notes.
\newblock \emph{Proceedings of SemEval}, pages 1292--1297, 2016.

\bibitem[Leeuwenberg and Moens(2016)]{leeuwenberg2016kuleuven}
Artuur Leeuwenberg and Marie-Francine Moens.
\newblock Kuleuven-liir at semeval 2016 task 12: Detecting narrative
  containment in clinical records.
\newblock \emph{Proceedings of SemEval}, pages 1280--1285, 2016.

\bibitem[Li et~al.(2011)Li, Dong, Maurino, and Srivastava]{li2011linking}
Pei Li, X~Dong, Andrea Maurino, and Divesh Srivastava.
\newblock Linking temporal records.
\newblock \emph{Proceedings of the VLDB Endowment}, 4\penalty0 (7):\penalty0
  956--967, 2011.

\bibitem[Madhavan et~al.(2007)Madhavan, Jeffery, Cohen, Dong, Ko, Yu, and
  Halevy]{32784}
Jayant Madhavan, Shawn~R. Jeffery, Shirley Cohen, Xin~(Luna) Dong, David Ko,
  Cong Yu, and Alon Halevy.
\newblock Web-scale data integration: You can only afford to pay as you go.
\newblock In \emph{CIDR}, 2007.

\bibitem[Movshovitz-Attias et~al.(2010)Movshovitz-Attias, Whang, Noy, and
  Halevy]{MovshovitzAttias2015}
Dana Movshovitz-Attias, Steven~Euijong Whang, Natalya Noy, and Alon Halevy.
\newblock Discovering subsumption relationships for web-based ontologies.
\newblock In \emph{Proceedings of the 18th International Workshop on Web and
  Databases}, WebDB'15, pages 62--69, New York, NY, USA, 2010. ACM.
\newblock ISBN 978-1-4503-3627-7.

\bibitem[Nebel and B{\"u}rckert(1995)]{nebel1995reasoning}
Bernhard Nebel and Hans-J{\"u}rgen B{\"u}rckert.
\newblock Reasoning about temporal relations: a maximal tractable subclass of
  {A}llen's interval algebra.
\newblock \emph{Journal of the ACM (JACM)}, 42\penalty0 (1):\penalty0 43--66,
  1995.

\bibitem[Nemhauser and Wolsey(1988)]{NemhauserWolsey:1985}
George~L. Nemhauser and Laurence~A. Wolsey.
\newblock \emph{Integer and combinatorial optimization}.
\newblock Wiley, New York;Chichester;, 1988.
\newblock ISBN 9780471828198;047182819X;.

\bibitem[Papadimitriou and Steiglitz(1998)]{papadimitriou1998combinatorial}
Christos~H Papadimitriou and Kenneth Steiglitz.
\newblock \emph{Combinatorial optimization: algorithms and complexity}.
\newblock Courier Corporation, 1998.

\bibitem[Powers(2011)]{Powers}
David M.~W. Powers.
\newblock Evaluation: from precision, recall and f-measure to roc,
  informedness, markedness and correlation.
\newblock \emph{Journal of Machine Learning Technologies}, 2\penalty0
  (1):\penalty0 37--63, 2011.

\bibitem[Punyakanok et~al.(2004)Punyakanok, Roth, Yih, and
  Zimak]{Punyakanok1346}
Vasin Punyakanok, Dan Roth, Wen-tau Yih, and Dav Zimak.
\newblock Semantic role labeling via integer linear programming inference.
\newblock In \emph{Proceedings of the 20th international conference on
  Computational Linguistics}, page 1346. Association for Computational
  Linguistics, 2004.

\bibitem[Pustejovsky and Stubbs(2011)]{pustejovsky2011increasing}
James Pustejovsky and Amber Stubbs.
\newblock Increasing informativeness in temporal annotation.
\newblock In \emph{Proceedings of the 5th Linguistic Annotation Workshop},
  pages 152--160. Association for Computational Linguistics, 2011.

\bibitem[Pustejovsky et~al.(2005)Pustejovsky, Ingria, Sauri, Castano, Littman,
  Gaizauskas, Setzer, Katz, and Mani]{Pustejovsky2005545}
James Pustejovsky, Bob Ingria, Roser Sauri, Jose Castano, Jessica Littman, Rob
  Gaizauskas, Andrea Setzer, Graham Katz, and Inderjeet Mani.
\newblock The specification language timeml.
\newblock \emph{The language of time: A reader}, pages 545--557, 2005.

\bibitem[Roth and Yih(2004)]{Roth2004}
Dan Roth and Wen-tau Yih.
\newblock A linear programming formulation for global inference in natural
  language tasks.
\newblock In \emph{Proceedings of CoNLL-2004}, 2004.

\bibitem[Saur\'{\i} et~al.(2005)Saur\'{\i}, Knippen, Verhagen, and
  Pustejovsky]{Sauri2005}
Roser Saur\'{\i}, Robert Knippen, Marc Verhagen, and James Pustejovsky.
\newblock Evita: A robust event recognizer for qa systems.
\newblock In \emph{Proceedings of the Conference on Human Language Technology
  and Empirical Methods in Natural Language Processing}, HLT '05, pages
  700--707, Stroudsburg, PA, USA, 2005. Association for Computational
  Linguistics.

\bibitem[Saur{\' i} et~al.(2009)Saur{\' i}, Goldberg, Verhagen, and
  Pustejovsky]{Sauri}
Roser Saur{\' i}, Lotus Goldberg, Marc Verhagen, and James Pustejovsky.
\newblock Annotating events in {E}nglish. {TimeML} annotation guidelines, 2009.
\newblock Brandeis University. Version TempEval-2010.

\bibitem[Sawilowsky(2009)]{sawilowsky2009new}
Shlomo~S Sawilowsky.
\newblock New effect size rules of thumb.
\newblock \emph{Journal of Modern Applied Statistical Methods}, 8:\penalty0
  597--599, 2009.

\bibitem[Schrijver(2003)]{schrijver2003combinatorial}
Alexander Schrijver.
\newblock \emph{Combinatorial optimization: polyhedra and efficiency},
  volume~24.
\newblock Springer Science \& Business Media, 2003.

\bibitem[Styler~IV et~al.(2014)Styler~IV, Bethard, Finan, Palmer, Pradhan,
  de~Groen, Erickson, Miller, Lin, Savova, et~al.]{styler2014temporal}
William~F Styler~IV, Steven Bethard, Sean Finan, Martha Palmer, Sameer Pradhan,
  Piet~C de~Groen, Brad Erickson, Timothy Miller, Chen Lin, Guergana Savova,
  et~al.
\newblock Temporal annotation in the clinical domain.
\newblock \emph{Transactions of the Association for Computational Linguistics},
  2:\penalty0 143--154, 2014.

\bibitem[UzZaman(2012)]{UzZaman2012Th}
Naushad UzZaman.
\newblock \emph{Interpreting the Temporal Aspects of Language}.
\newblock Thesis, University of Rochester, 2012.

\bibitem[UzZaman et~al.(2013)UzZaman, Llorens, Allen, Derczynski, Verhagen, and
  Pustejovsky]{UzZaman2012}
Naushad UzZaman, Hector Llorens, James Allen, Leon Derczynski, Marc Verhagen,
  and James Pustejovsky.
\newblock Semeval-2013 task 1: Tempeval-3: Evaluating time expressions, events,
  and temporal relations.
\newblock In \emph{Second Joint Conference on Lexical and Computational
  Semantics (* SEM)}, pages 1--9. Association for Computational Linguistics,
  2013.
\newblock Also see preprint arXiv:1206.5333.

\bibitem[Verhagen et~al.(2009)Verhagen, Gaizauskas, Schilder, Hepple,
  Moszkowicz, and Pustejovsky]{Verhagen}
Marc Verhagen, Robert Gaizauskas, Frank Schilder, Mark Hepple, Jessica
  Moszkowicz, and James Pustejovsky.
\newblock The tempeval challenge: Identifying temporal relations in text.
\newblock \emph{Language Resources and Evaluation}, 43\penalty0 (2):\penalty0
  161--179, 2009.
\newblock ISSN 1574020X, 15728412.
\newblock URL \url{http://www.jstor.org/stable/27743609}.

\bibitem[Woodsend and Lapata(2011)]{Woodsend}
Kristian Woodsend and Mirella Lapata.
\newblock Learning to simplify sentences with quasi-synchronous grammar and
  integer programming.
\newblock In \emph{Proceedings of the 2011 Conference on Empirical Methods in
  Natural Language Processing}, pages 409--420. Association for Computational
  Linguistics, 2011.

\end{thebibliography}

\end{document}